%% file: main.tex
\documentclass[times]{acsauth}
\usepackage{hyperref}
\usepackage{blindtext}
\usepackage{graphicx}
\usepackage{epstopdf}
\usepackage{float}
\usepackage{setspace}

\usepackage{mathrsfs}
\graphicspath{ {./images/} }
\usepackage{amssymb}
\usepackage{caption}
\usepackage{subcaption}
\usepackage[noadjust]{cite}
\usepackage{multirow}
\usepackage{array}
\usepackage{inputenc}
\usepackage[inline]{enumitem}
\usepackage{soul}
\usepackage{booktabs}
\usepackage{varwidth}
\usepackage{bm}

\usepackage{amsthm}
\usepackage{tikz}

\theoremstyle{plain}
\newtheorem{theorem}{{Theorem}}
\newtheorem{lemma}{{Lemma}}
\newtheorem{assumption}{{Assumption}}
\newtheorem{remark}{{Remark}}

\newcommand{\bbm}{\begin{bmatrix}}
\newcommand{\ebm}{\end{bmatrix}}

\newcommand{\lone}{{\mathcal{L}_1}}
\newcommand{\linf}{{\mathcal{L}_\infty}}

\newcommand{\bff}[2]{{\bf{#1}}_{#2}}
\newcommand{\hbff}[2]{\widehat{\bf{#1}}_{#2}}
\newcommand{\bbff}[2]{\bar{\bf{#1}}_{#2}}

\newcommand{\rev}[1]{#1}
\newcommand{\lnone}[1]{\left\Vert#1\right\Vert_{\lone}}
\newcommand{\lninf}[1]{\left\Vert#1\right\Vert_{\linf}}
\renewcommand\arraystretch{1.6}

\setlist*[enumerate,1]{%
  label=(\roman*),
}

\usepackage{moreverb}

\DeclareCaptionLabelFormat{andtable}{#1~#2  \&  \tablename~\thetable}

\newcommand\BibTeX{{\rmfamily B\kern-.05em \textsc{i\kern-.025em b}\kern-.08em
T\kern-.1667em\lower.7ex\hbox{E}\kern-.125emX}}

\def\volumeyear{2018}

\begin{document}

\runningheads{Transfer Learning for High-Precision Trajectory Tracking}{K. Pereida \textit{Et Al.}}

\title{Transfer Learning for High-Precision Trajectory Tracking Through $\lone$ Adaptive Feedback and Iterative Learning}

\author{Karime~Pereida,~Dave~Kooijman,~Rikky~R.~P.~R.~Duivenvoorden,~and~Angela~P.~Schoellig\corrauth}

\address{Institute for Aerospace Studies, University of Toronto, North York, ON M3H 5T6, Canada}

\corraddr{Karime Pereida, Institute for Aerospace Studies, Univeristy of Toronto, 4925 Dufferin St, North York, ON M3H 5T6, Canada.}

\cgs{Canada Foundation for Innovation John R. Evans Leaders Fund, Grant/Award Number: CFI/ORF 33000; Natural Sciences and Engineering Research Council of Canada, Grant/Award Number: CREATE-466088, RGPIN-2014-04634; Ontario Research Fund for Small Infrastructure Funds CFI/ORF 33000; Alfred P. Sloan Foundation Sloan Research Fellowship and Ontario Early Researcher Award; Mexican National Council of Science and Technology}

\begin{abstract}
Robust and adaptive control strategies are needed when robots or automated systems are introduced to unknown and dynamic environments, where they are required to cope with disturbances, unmodeled dynamics and parametric uncertainties. In this paper, we demonstrate the capabilities of a combined $\lone$ adaptive control and iterative learning control (ILC) framework to achieve high-accuracy trajectory tracking in the presence of unknown and changing disturbances. The $\lone$ adaptive controller makes the system behave close to a reference model; however, it does not guarantee that perfect trajectory tracking is achieved, while ILC improves trajectory tracking performance based on previous iterations. The combined framework in this paper uses $\lone$ adaptive control as an underlying controller that achieves a robust and repeatable behavior, while the ILC acts as a high-level adaptation scheme that mainly compensates for systematic tracking errors. We illustrate that this framework enables transfer learning between dynamically different systems, where learned experience of one system can be shown to be beneficial for another, different system. Experimental results with two different quadrotors show the superior performance of the combined $\lone$-ILC framework compared to approaches using ILC with an underlying proportional-derivative (PD) or proportional-integral-derivative (PID) controller. Results highlight that our $\lone$-ILC framework can achieve high-accuracy trajectory tracking when unknown and changing disturbances are present and can achieve transfer of learned experience between dynamically different systems. Moreover, our approach is able to achieve accurate trajectory tracking in the first attempt when the initial input is generated based on the reference model of the adaptive controller.
\end{abstract}


\maketitle
\vspace{-6pt}
\section{Introduction}
\vspace{-2pt}
\input{Introduction.tex}

\section{Problem statement}
\label{sec:problem}
\input{ProblemStatement.tex}

\section{Methodology}
\label{sec:methodology}
\input{Methodology.tex}

\section{Experimental Results}
\label{sec:results}
\input{Experiments.tex}

\section{Conclusion}
\label{sec:conclusion}
\input{Conclusion.tex}


\appendix

\section{Proof of Lemma~\ref{lem:BIBOref}} \label{app:lemBIBO}
\input{Appendix.tex}

\section{Proof of Theorem~\ref{th:boundness}} \label{app:thboundness}
\input{Appendix2.tex}

\section{Discussion of Remark~\ref{rem:convergence}} \label{app:remconvergence}
\input{Appendix3.tex}


\bibliographystyle{wileyj}
\bibliography{journalbib}
\end{document}

%% file: Introduction.tex
\begin{figure}[t]
\centering
\includegraphics[width=0.8\linewidth]{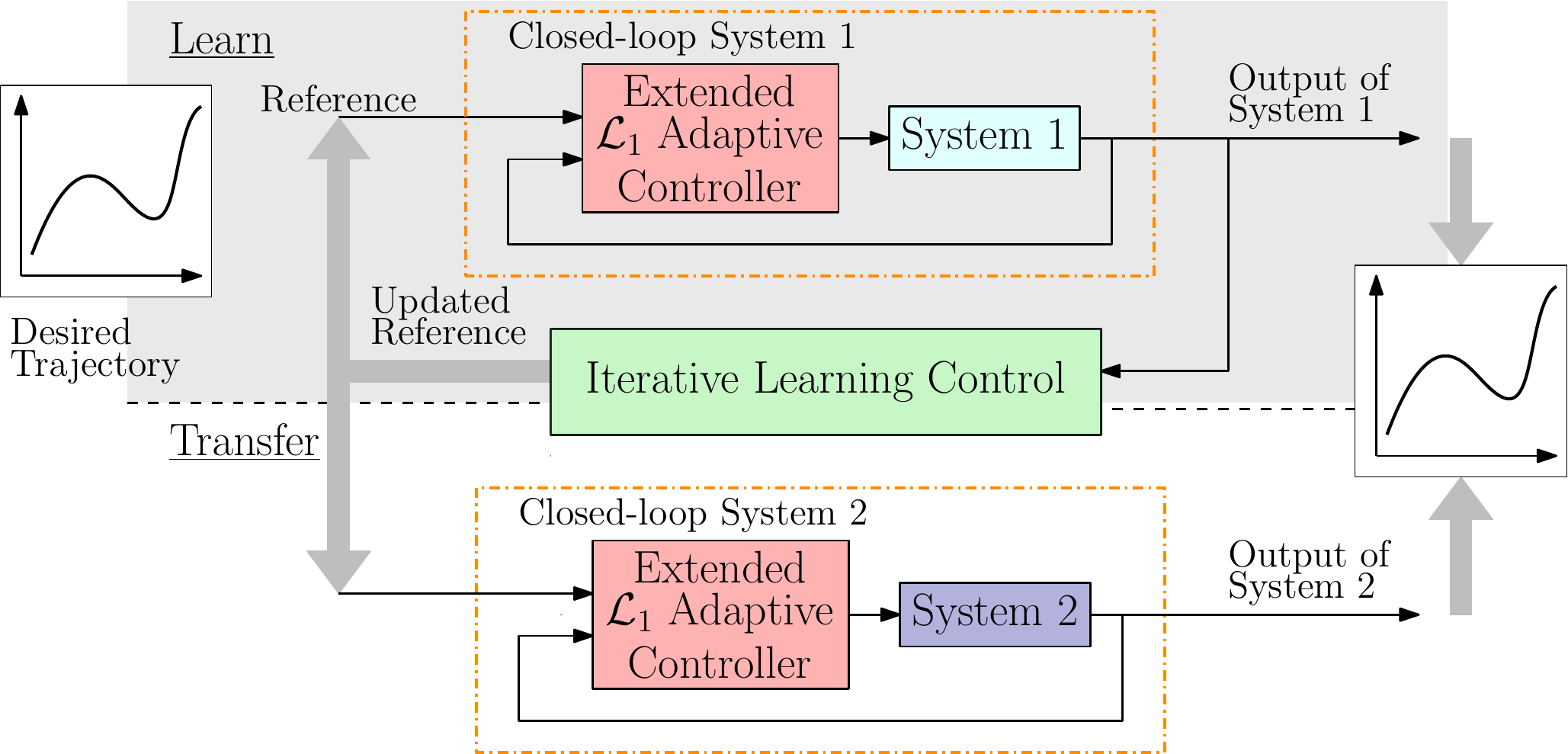}
\caption{The Extended $\lone$ Adaptive Controller forces a system to behave in a repeatable, predefined way. Consequently, two dynamically different systems (System 1 and System 2) can achieve the same predefined behavior, noted by orange dashed boxes. Extended $\lone$ Adaptive Controller boxes share the same color as they have the same reference system. The Iterative Learning Controller is capable of learning an input such that the output tracks a desired output signal. After learning a trajectory with System 1, the learned input can be applied to other systems with different dynamical behavior, such as System 2, and the output will still track the reference signal.}
\label{fig:transfer_learning}
\end{figure}

Robots and automated systems are being deployed in unstructured and continuously changing environments. Sophisticated control methods are required to guarantee high overall performance in these environments where model uncertainties, unknown disturbances and changing dynamics are present. Examples of robotic applications in unknown, dynamic environments include autonomous driving, assistive robotics and unmanned aerial vehicle (UAV) applications. In applications where expensive hardware is involved, it is advantageous to use simulators or inexpensive hardware for the initial control design and learning. However, learned trajectories in the inexpensive hardware should be transferred, without further processing, to a different system and achieve a performance comparable to the one obtained in the training system (see Fig.~\ref{fig:transfer_learning}). Moreover, to achieve high trajectory tracking performance the underlying controller must be robust enough as small changes in the conditions may otherwise result in a dramatic decrease in controller performance and could cause instability (see \cite{Skelton1989}, \cite{Morari1999} and \cite{Skogestad2007}).

The objective of this paper is to design a framework that makes the system achieve a repeatable behavior even in the presence of unknown disturbances and changing dynamics, that improves performance over time and that is able to transfer learned trajectories to dynamically different systems achieving high-accuracy trajectory tracking. Therefore we propose a combined $\lone$ adaptive control and iterative learning control (ILC) framework (see Fig.~\ref{fig:transfer_learning}). 

Control frameworks that combine the advantages of repeatable behavior and improved performance over time have been proposed. In particular, we focus on a combined adaptive control and learning control framework. Adaptive control methods deal with model uncertainties and unknown disturbances. Model reference adaptive control (MRAC) uses the difference between the output of the system and the output of a desired reference model to update control parameters. The goal is for the parameters to converge so the plant response matches the reference model response \cite{Parks1966}. Large adaptive gains help to achieve this goal; however, they result in high-frequency oscillations in the control signal \cite{Hovakimyan2010}. The $\lone$ adaptive controller is based on the MRAC architecture with the addition of a low-pass filter that decouples robustness from adaptation. This allows arbitrarily high adaptation gains to be chosen for fast adaptation and to determine uniform bounds for the system's state and control signals \cite{Hovakimyan2010}. Attitude control based on $\lone$ adaptive control was shown in \cite{Mallikarjunan2012}, where three algorithms were successfully implemented and tested on a quadrotor, hexacopter and octocopter, respectively. $\lone$ adaptive output feedback on translational velocity was successfully implemented on a quadrotor to compensate for artificial reduction in the speed of a single motor \cite{Michini2009}.
 
Iterative learning control (ILC) is used to efficiently calculate the feedforward input signal by using information from previous trials to improve tracking performance in a small number of iterations. ILC has been successfully applied to a variety of trajectory tracking scenarios such as robotic arms \cite{Gunnarsson2001}, ground vehicles \cite{Ostafew2013}, manufacturing of integrated circuits \cite{Yu2014}, swinging up a pendulum  \cite{Schoellig2009}, and quadrotor control \cite{Mueller2012}. An ILC method based on minimization of a quadratic performance criterion was used in \cite{Schoellig2012} for precise quadrocopter trajectory tracking. A survey on ILC can be found in \cite{Bristow2006}.

A framework of $\lone$ adaptive feedback control with parallel ILC was proposed in \cite{Barton2011}, \cite{Altin2013}, and \cite{Altin2014} where successful simulation results were presented. In the parallel framework the input to the system is the addition of the input signal calculated by the $\lone$ adaptive controller and the input signal calculated by the ILC. The addition of the two input signals couples the problems of providing a repeatable system behavior with the problem of improving tracking performance. In \cite{Pereida2017} we proposed and showed the first experimental results on a quadrotor of a serial framework of $\lone$ adaptive control and ILC under changing dynamics. The serial architecture uses the $\lone$ adaptive controller as an underlying control to achieve a repeatable system behavior despite the presence of unknown and changing disturbances. It then applies ILC to the now repeatable system to improve trajectory tracking performance. In this work we exploit the repeatable behavior of the $\lone$ adaptive controller to transfer learning between dynamically different systems while achieving high-accuracy trajectory tracking.  

Different strategies for transferring learning data from simulation to the real world have previously been proposed. In~\cite{Tobin2017}, simulated images generated by randomizing rendering in a simulator are used to train models for object localization. These models transfer to real images and are accurate enough to be used to perform grasping in cluttered environments. Furthermore, noting that policies that succeed in simulation often do not work when deployed in a real robot, \cite{Christiano2016} proposed to use what a simulation-based control policy expects the next state(s) will be and, based on a learned deep inverse dynamics model, calculate which real-world action is most suitable to achieve those states. However, to accurately train a deep inverse dynamics model a significant amount of real data is required. In contrast, our work only requires both systems to behave in the same predefined way, through the use of an adaptive controller, to be able to transfer learning from simulation to real world. Moreover, a strategy that transfers learning from specific skills and robots to different skills and robots was proposed in \cite{Devin2016}. To achieve this, ``task-specific" and ``robot-specific" neural network policies are composed and then trained end-to-end. When an unseen combination is encountered, the appropriate ``task-specific" and ``robot-specific" previously trained policies are composed to solve the new robot-task combination. In our work we focus on achieving the same behavior with different systems, so we are not required to relearn for each different system. 
 
In this work we show the capabilities of the combined $\lone$ adaptive control and ILC framework to achieve high-accuracy trajectory tracking even if 
 \begin{enumerate*}[font=\itshape] 
 \item changing system dynamics and uncertain environment conditions are present, and 
 \item learned trajectories are transferred between dynamically different systems.
 \end{enumerate*}
We also show that a reference trajectory generated based on the reference model of the adaptive controller achieves more accurate tracking performance than reference trajectories generated with standard choices. We use the serial framework \cite{Pereida2017} where the $\lone$ adaptive controller acts as an underlying controller (see Fig.~\ref{fig:blockdiagram}) that makes the system display a repeatable and reliable behavior (in other words, it achieves the same output when the same reference input is applied) even in the presence of unknown disturbances and changing dynamics; however, perfect trajectory tracking is not achieved. After each iteration the ILC improves the tracking performance of the now repeatable system using knowledge from previous iterations. 

The $\lone$ adaptive controller forces systems to follow a predefined behavior defined through a so-called reference model, even if the systems are dynamically different. Therefore, learned trajectories in one system can be transferred among dynamically different systems (that have an underlying $\lone$ adaptive controller with the same reference model) to achieve perfect tracking or to significantly decrease the initial tracking error in a different system (see Fig.~\ref{fig:transfer_learning}). Experimental results on two dynamically different quadrotors show that the proposed approach achieves high trajectory tracking performance despite the presence of unknown disturbances. Furthermore, we show that our approach allows us to train on a simulator or on a quadrotor then transfer the learned trajectory to a dynamically different quadrotor and achieve a high-accuracy tracking performance even in the first iteration. The tracking performance achieved by our approach cannot, under changing dynamics, be achieved by baseline proportional-derivative (PD) and proportional-derivative-integral (PID) controllers combined with ILC. However, ILC has the limitation of not being able to generalize previously learned tasks to new, unseen tasks. In future work a linear map generated using prior knowledge from previously learned trajectories (see \cite{Hamer2013}) could be used to achieve transfer learning between different robots and tasks. 
 
The remainder of this paper is organized as follows: we define the problem in Section~\ref{sec:problem}. The details of the proposed approach and proofs of key features are presented in Section~\ref{sec:methodology}. Section~\ref{sec:results} shows our experimental results, including examples where learned trajectories are transferred between dynamically different systems. We compare our approach to two frameworks with  standard underlying feedback controllers. Conclusions are provided in Section~\ref{sec:conclusion}.

%% file: ProblemStatement.tex
The objectives of this work are to achieve high-accuracy trajectory tracking %
\begin{enumerate*}[font=\itshape] 
  \item{when changing system dynamics and uncertain environment conditions are present, } 
  \item {in a new and dynamically different system by transferring the previously learned trajectories, and } 
  \item {by calculating the initial reference input based on the model reference defined in the $\lone$ adaptive controller . }
\end{enumerate*}
For a given desired trajectory the system optimizes its performance over multiple executions and, if required, transfers the learned trajectory to a dynamically different system that is able to achieve a similar, optimized performance. Moreover, even if the system dynamics continue to change, there is no need to re-learn. 

\begin{figure}[t]
\centering
\centering{
\includegraphics[width=0.8\textwidth]{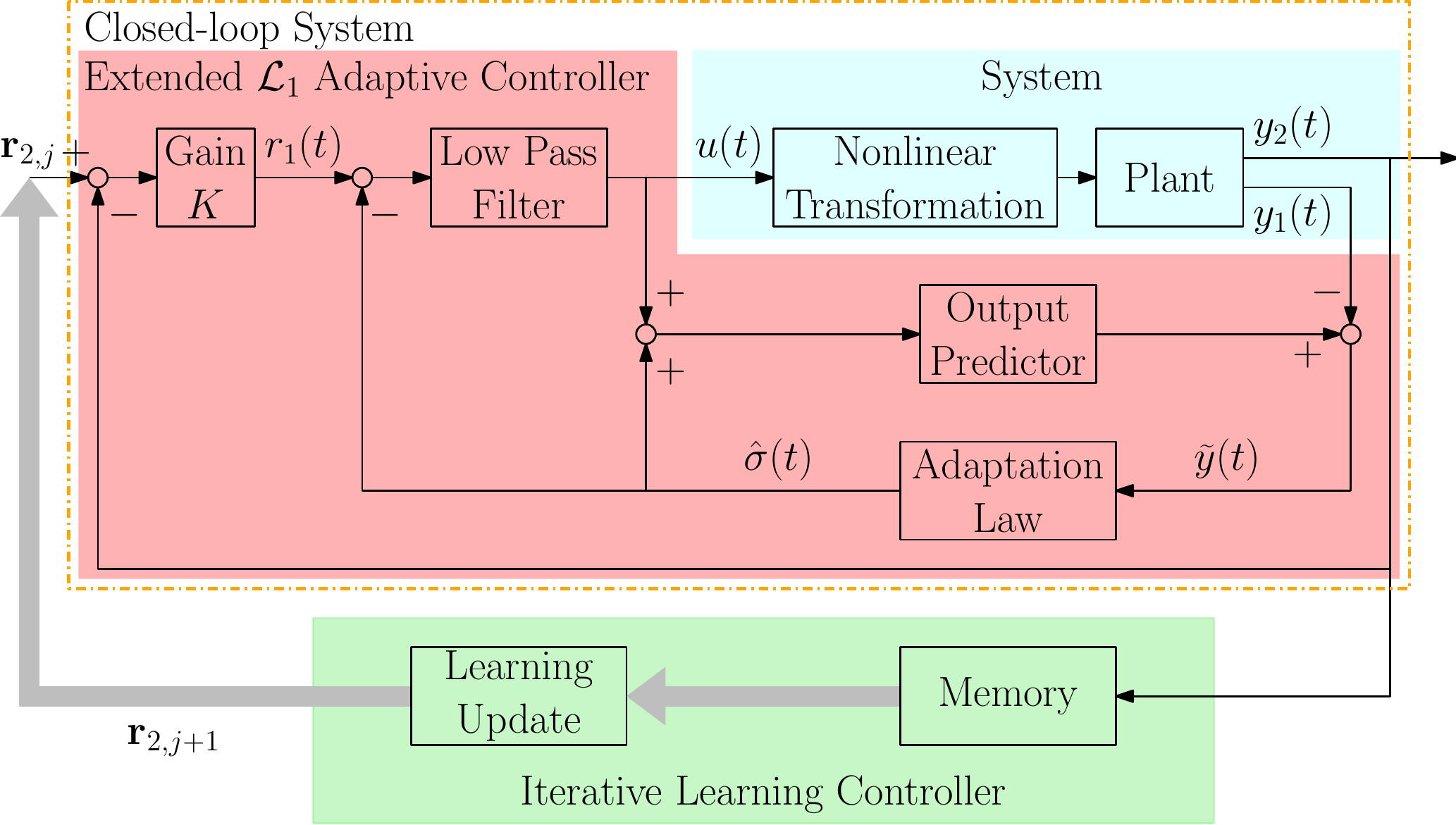}
}
\vspace{-0.3cm}
\caption{Proposed framework to achieve high-accuracy trajectory tracking in changing environments. The extended $\lone$ adaptive controller forces the system to behave in a predefined, repeatable way. The iterative learning controller improves the tracking performance in each iteration $j$ based on experience from previous executions.} 
\label{fig:blockdiagram}
\end{figure}

We assume that the uncertain and changing dynamics (`System' block in Fig.~\ref{fig:blockdiagram}) can be described by a single-input single-output (SISO) system (this approach can be extended to multi-input multi-output (MIMO) systems as described in Section~\ref{sssec:MIMO}) identical to \cite{Hovakimyan2010} for output feedback: 
\begin{equation}
\label{eq:sisosystem}
\begin{array}{l l}
  y_1(s) = A(s) (u(s)+d_{\lone}(s)), & y_2(s)=\frac{1}{s}y_1(s),
\end{array} 
\end{equation}
where $y_1(s)$ and $y_2(s)$ are the Laplace transforms of the translational velocity $y_1(t)$, and position $y_2(t)$, respectively, $A(s)$ is a strictly-proper \emph{unknown} transfer function that can be stabilized by a proportional-integral controller, $u(s)$ is the Laplace transform of the input signal, and $d_{\lone}(s)$ is the Laplace transform of the disturbance signal defined as $d_{\lone}(t)\triangleq f(t,y_1(t))$, where $f:\mathbb{R}\times \mathbb{R}\rightarrow \mathbb{R}$ is an \emph{unknown} map subject to the assumption: 

\begin{assumption}[Global Lipschitz continuity]
\label{asm:globallip}
There exist constants $L>0$ and $L_0>0$, such that the following inequalities hold uniformly in $t$:
\[
\begin{array}{r c l}
|f(t,v)-f(t,w)| & \leq & L|v-w|,\ and \\
|f(t,w)| & \leq & L|w| + L_0 \quad \forall v,w\in\mathbb{R}.
\end{array}
\]
\end{assumption}

The system is tasked to track a desired position trajectory $y_2^*(t)$, which is defined over a finite-time interval and is assumed to be feasible with respect to the true dynamics of the $\lone$-controlled system (red and blue boxes in Fig.~\ref{fig:blockdiagram}). This signal is discretized because the input of computer-controlled systems is sampled and measurements are only available at fixed time intervals. We introduce the lifted representation, see \cite{Gunnarsson2001}, for the desired trajectory ${\bf{y}}^*_2 = (y^*_2(1),\hdots,y^*_2(N))$, the output of the plant ${\bf{y}}_2 = (y_2(1),\hdots,y_2(N))$, and the reference input ${\bf{r}}_2 = (r_2(1),\hdots,r_2(N))$, where $N<\infty$ is the number of discrete samples. The tracking performance criterion $J$ is defined as: 
\begin{equation}
J\triangleq \min_{\bf{e}} {\bf{e}}^T{\bf{Q}}{\bf{e}}
\label{eq:costfunctione}
\end{equation}
where ${\bf{e}}={\bf{y}}_2-{\bf{y}}_2^*$ is the tracking error and ${\bf{Q}}$ is a positive-definite matrix. In this way the reference input ${\bf{r}}_2$ is updated to improve the trajectory tracking iteratively.

%% file: Methodology.tex
We consider two main subsystems: the extended $\lone$ adaptive controller (red box in Fig.~\ref{fig:blockdiagram}) and the ILC (green box in Fig.~\ref{fig:blockdiagram}). The extended $\lone$ adaptive controller is presented in Section~\ref{ssec:lone} including proofs on its transient behavior when subjected to (dynamic) disturbances. Section~\ref{ssec:ilc} introduces the ILC and includes a remark on convergence. \rev{Section~\ref{ssec:transfer} discusses the transfer of learned trajectories between dynamically different systems.}

\subsection{\texorpdfstring{$\lone$}{L1} Adaptive Feedback}
\label{ssec:lone}
The goal of the $\lone$ adaptive controller in this framework is to make the system behave in a repeatable, predefined way, even when unknown and changing disturbances affect the system. A description of the extended $\lone$ adaptive controller and transient behavior proofs are presented next. 

The extended architecture used in this work is identical to \cite{Michini2009}, where the typical $\lone$ adaptive output feedback controller for SISO systems acting on velocity \cite{Hovakimyan2010} is nested within a proportional controller (see Fig.~\ref{fig:blockdiagram}). The outer-loop proportional controller enables the system to remain within certain position boundaries. 

\subsubsection{Problem Formulation:}
The $\lone$ adaptive output feedback controller aims to design a control input $u(t)$ such that the output $y_2(t)$ tracks a bounded piecewise continuous reference input $r_2(t)$. We aim to achieve a desired closed-loop behavior, where the output of the $\lone$ adaptive controller $y_1(t)$, nested within a proportional feedback loop, tracks $r_1(t)$ according to a first-order reference dynamic system: 
\begin{equation}
\label{eq:referencesystem}
M(s) = \frac{m}{s+m}\, ,\quad m>0. 
\end{equation}

\subsubsection{Definitions and \texorpdfstring{$\lone$}{L1}-Norm Condition:}
The system in \eqref{eq:sisosystem} can be rewritten in terms of the reference system \eqref{eq:referencesystem}: 
\begin{equation}
y_1(s)=M(s)(u(s)+\sigma(s)),
\label{eq:newsystem}
\end{equation}
where uncertainties in $A(s)$ and $d_{\lone}(s)$ are combined into $\sigma$: 
\begin{equation}
\sigma(s)\triangleq\frac{(A(s)-M(s))u(s)+A(s)d_{\lone}(s)}{M(s)}.
\label{eq:defsigma}
\end{equation}
We consider a strictly-proper low-pass filter $C(s)$ (see Fig.~\ref{fig:blockdiagram}) with $C(0)=1$, and a proportional gain $K\in\mathbb{R}^+$, such that: 
\begin{equation}
H(s)\triangleq \frac{A(s)M(s)}{C(s)A(s)+(1-C(s))M(s)} \quad \text{is stable,}
\label{eq:defH}
\end{equation}
\begin{equation}
F(s)\triangleq \frac{1}{s+H(s)C(s)K} \quad \text{is stable,}
\label{eq:defF}
\end{equation}
and the following $\lone$-norm condition is satisfied:\rev{
\begin{equation}
\lnone{G(s)}L < 1\, ,\ \ \text{ where }\ \  G(s)\triangleq H(s)(1-C(s))\,,
\label{eq:l1normcondition}
\end{equation} }
where $L$ is the Lipschitz constant defined in Assumption~\ref{asm:globallip}. The transfer function $H(s)$ helps to describe the relationship between $y_1(s)$ and $r_1(s)$ and between $y_1(s)$ and $d_\lone(s)$. It is obtained from equations (4.90)-(4.92) in \cite{Hovakimyan2010}. The transfer function $F(s)$ helps to describe the relationship between $y_2(s)$ and $r_2(s)$ and between $y_2(s)$ and $d_\lone(s)$. To obtain $F(s)$, we substitute $r_1(s) = K (r_2(s)-y_2(s))$ into equation (4.94) in \cite{Hovakimyan2010} and use the result to solve for $y_2(s)$ in \eqref{eq:sisosystem}. Finally, the transfer function $G(s)$ describes the relationship between $y_2(s)$ and $d_\lone(s)$.

To prove the bounded-input bounded-output (BIBO) stability of a reference model, which describes the repeatable behavior of the extended $\lone$ controlled system, the $\lone$-norm condition is used. The solution of the $\lone$-norm condition in \eqref{eq:l1normcondition} exists under the following assumptions: 
\begin{assumption}[Stability of H(s)]
The transfer function H(s) is assumed to be stable for appropriately chosen low-pass filter $C(s)$ and first-order reference eigenvalue $-m<0$.
\end{assumption}

This assumption holds when $A(s)$ can be stabilized by a proportional-integral controller \cite{Hovakimyan2010}. 

\begin{assumption}[Stability of F(s)]
The transfer function F(s) is assumed to be stable for appropriately chosen proportional gain $K$.
\end{assumption}

For this assumption to be valid, a sufficient condition is that $A(s)$ is minimum-phase stable, which holds if there is a controller within the system $A(s)$ that is stabilizing the plant without any unstable zeros. This assumption is valid in the case of velocity control of a quadrotor. 

\subsubsection{Extended \texorpdfstring{$\lone$}{L1} Adaptive Control Architecture:}
The SISO extended $\lone$ adaptive controller architecture is shown in Fig.~\ref{fig:blockdiagram}. This architecture (from $r_1$ to $y_1$) is identical to \cite{Hovakimyan2010} with the exception of the proportional feedback loop. The integrator from $y_1$ to $y_2$ allows the outer-loop to control position, while the $\lone$ adaptive feedback controls the velocity. The equations that describe the implementation of the extended $\lone$ output feedback architecture are presented below.
\begin{description}
\item [Output Predictor:] The output predictor used within the $\lone$ adaptive output feedback architecture is: 
\[
\dot{\hat{y}}_1(t)=-m\hat{y}_1(t)+m(u(t)+\hat{\sigma}(t))\, , \quad \hat{y}_1(0)=0\, ,
\]
where $\hat{\sigma}(t)$ is the adaptive estimate of $\sigma(t)$. In the Laplace domain, this is: 
\begin{equation}
\hat{y}_1(s)=M(s)(u(s)+\hat{\sigma}(s)).
\label{eq:outputpredictor}
\end{equation}
\item[Adaptation Law:] The adaptive estimate $\hat{\sigma}$ is updated according to the following update law: 
\begin{equation}
\dot{\hat{\sigma}}(t)=\Gamma \text{Proj} (\hat{\sigma}(t),-mP\tilde{y}(t))\, , \quad \hat{\sigma}(0)=0\,,
\label{eq:adaptationlaw}
\end{equation}
where $\tilde{y}(t)\triangleq \hat{y}_1(t)-y_1(t)$, and $P>0$ solves the algebraic Lyapunov equation $mP+Pm=2mP=-Z$ for $Z>0$. The adaptation rate $\Gamma\in\mathbb{R}^+$ is subject to the lower bound specified in \cite{Hovakimyan2010}. For a fast adaptation, $\Gamma$ is set very large. The projection operator is defined in \cite{Hovakimyan2010} and ensures that the estimation of $\sigma$ is guaranteed to remain within a specified convex set which contains all possible values of $d_\lone(s)$ and the range of uncertainties in $A(s)$. Intuitively, this convex set includes all the values that $\sigma$ in \eqref{eq:newsystem} could take. 
\item[Control Law:] The control input is a low-pass filtered signal by $C(s)$ of the difference between the $\lone$ desired trajectory $r_1$ and the adaptive estimate $\hat{\sigma}$: 
\begin{equation}
u(s)=C(s)(r_1(s)-\hat{\sigma}(s))\,.
\label{eq:controllaw}
\end{equation}
Hence, it only compensates for the low frequencies of the uncertainties within $A(s)$ and $d_{\lone}$, which the system is capable of counteracting. The high-frequency portion is attenuated by the low-pass filter. 
\item[Closed-Loop Feedback:] The objective of the closed-loop feedback is for $y_2$ to track $r_2$. It acts on the input to the $\lone$ adaptive output feedback controller $r_1$ based on the output of the system $y_1$. From above, $y_2(s)\triangleq \frac{1}{s}y_1(s)$, and the negative feedback is defined as: 
\begin{equation}
r_1(s)=K(r_2(s)-y_2(s))\,. 
\label{eq:clfeedback}
\end{equation}
\end{description}

\subsubsection{Transient and Steady-State Performance:} 
The extended $\lone$ adaptive controller guarantees that the difference between the output of a given BIBO stable reference system and the output of the actual system is uniformly bounded. In other words, the actual system behaves close to the reference system. Intuitively, the extended $\lone$ adaptive controller makes the system perform repeatably and consistently. 

\rev{We first introduce the following assumption necessary to prove uniform boundedness of the difference between the output of a given BIBO stable reference system and the output of the actual system.}

\rev{
\begin{assumption}[Boundedness of $r_1(t)$]
 The signal $r_1(t)$ is assumed to be a bounded, piecewise continuous signal. Therefore, it has a bounded norm $\lninf{r_1}$.
\end{assumption}
}

\rev{The above assumption is justifiable since the $\lone$ adaptive controller makes the system behave close to the linear reference system $M(s)$. In other words, the low-pass filter, output predictor, adaptation law and system (see Fig.~\ref{fig:blockdiagram}) behave close to the linear model $M(s)$. By choosing $\text{det}(M(0))>0$, and using Theorem 1 in~\cite{Morari1985}, we know that there exists a $K=kI$, with $k>0$, such that the closed-loop system from $r_2$ to $y_2$ is stable; hence, $r_1(t)$ is bounded. The proof is part of future work as stability of integral controllers for nonlinear systems is an ongoing research topic (see, for example \cite{Konstantopoulos2016}) and outside the scope of the present work.}

We then present the BIBO stable closed-loop reference system.

\begin{lemma}
\rev{
Let $C(S)$, $M(S)$ and $K$ satisfy the $\lone$-norm condition in \eqref{eq:l1normcondition}. Then the following closed-loop reference system, 
\begin{eqnarray}
y_{1,ref}(s) & = & M(s)(u_{1,ref}(s)+\sigma_{ref}(s))\,, \label{eq:y1refM} \\
u_{1,ref}(s) & = & C(s)(r_{1,ref}(s)-\sigma_{ref}(s))\,, \label{eq:u1ref}\\
y_{2,ref}(s) & = & \frac{1}{s}y_{1,ref}(s)\,, \label{eq:y2refs} \\
r_{1,ref}(s) & = & K(r_2(s)-y_{2,ref}(s))\,, \label{eq:r1ref}
\end{eqnarray} }
\rev{where }
\rev{
\begin{equation}
 \sigma_{ref}(s) = \frac{(A(s)-M(s))u_{1,ref}(s)+A(s)d_{1,ref}(s)}{M(s)}\,,
 \label{eq:sigmadef}
\end{equation}
and $d_{1,ref}(s)$ is the Laplace transform of $d_{1,ref}(t) \triangleq  f(t,y_{1,ref}(t))$ is BIBO stable. 
}
\label{lem:BIBOref}
\end{lemma}

\rev{The proof of this lemma is found in Appendix~\ref{app:lemBIBO}. Next, we show that error in the estimation is bounded and that the system behaves close to the BIBO stable reference system. } 

\begin{theorem}
Consider the system in \eqref{eq:sisosystem}, with a control input from the extended $\lone$ output feedback adaptive controller defined in \eqref{eq:outputpredictor}, \eqref{eq:adaptationlaw}, \eqref{eq:controllaw} and \eqref{eq:clfeedback}. Suppose $C(s)$, $M(s)$ and $K$ satisfy the $\lone$-norm condition in \eqref{eq:l1normcondition}. Then the following bounds hold: 
\begin{align}
\Vert \tilde{y}\Vert_\linf  & \leq  \gamma_0\,, \label{eq:ytildebound} \\
\Vert y_{2,ref}-y_2\Vert_\linf & \leq  \gamma_1\, \label{eq:overallbound},
\end{align}
where $\tilde{y}(t)\triangleq\hat{y}_1(t)-y_1(t)$, $\gamma_0 \propto \sqrt{\frac{1}{\Gamma}}$ is defined in \cite{Hovakimyan2010}, and 
\rev{
\begin{equation}
\gamma_1 \triangleq \left( \lnone{F(s)G(s)}L\dfrac{\lnone{H_2(s)}}{1-\lnone{G(s)}L} + \lnone{\dfrac{F(s) H(s) C(s)}{M(s)} }\right)\gamma_0\,.
\end{equation}
\label{th:boundness}
}
\end{theorem}
\rev{The proof of this theorem is found in Appendix~\ref{app:thboundness}.}

The difference between the output predictor and the system output $y_1(t)$ and the difference between the reference system and the system output $y_2(t)$ are uniformly bounded with bounds inversely proportional to the square root of the adaptation gain $\Gamma$. For high adaptation gains, the actual system approaches the behavior of the reference system~\eqref{eq:y2refFd}. Hence, the system achieves repeatable and consistent performance, which is required for ILC.

\subsubsection{Multi-Input Multi-Output Implementation:}
\label{sssec:MIMO}
The SISO architecture derived so far can be extended to a multi-input multi-output (MIMO) implementation. In our application, it can be assumed that states are decoupled (after applying an appropriate feedback linearization). Hence, for $n$ different states, the low-pass filter $C(s)$ and the first-order output predictor \eqref{eq:outputpredictor} are implemented as $(n \times n)$ diagonal transfer function matrices: 
\[
\begin{array}{lr}
C(s) = \text{diag } (C_1(s),\hdots,C_n(s))\,,
& M(s) = \text{diag }(M_1(s),\hdots,M_n(s))\,,
\end{array}
\]
where $C_i(s)=\frac{\omega_i}{s+\omega_i}$, $M_i(s)=\frac{m_i}{s+m_i}$ and $i=1,\hdots,n$. Moreover, the proportional gain $K$ is implemented as an $(n\times n)$ matrix: 
\[
K = \text{diag }(k_1,\hdots,k_n)\,,
\]
where $k_i\in\mathbb{R}^+$.

\subsection{Iterative Learning Control}
\label{ssec:ilc}
In this work, we use the extended $\lone$ adaptive controller to achieve a repeatable system, even in the presence of disturbances, and ILC to improve tracking performance of the resulting repeatable system. We assume we have an approximate model of the repeatable system (orange dashed line in Fig.~\ref{fig:blockdiagram}): 
\begin{equation}
\dot{x}(t)=f(x(t),r_2(t)) \quad y_2(t)=h(x(t))\, ,
\label{eq:nominal}
\end{equation}
where $r_2(t)\in \mathbb{R}$ is the control input, $x(t)\in \mathbb{R}^{n_x}$ is the state, and $y_2(t)\in\mathbb{R}$ is the output. In order to be consistent with the assumptions made in Section~\ref{ssec:lone}, we have $r_2(t)$, $y_2(t)\in\mathbb{R}$; however, the approach described in this section can be extended to MIMO systems as described in Section~\ref{sssec:MIMOILC}. We assume that the system states can be directly measured or observed from the output. In many control applications, constraints must be placed on the process variables to ensure safe and smooth operations. The system may be subjected to input or output constraints of the form: 
\begin{equation}
V_{c}y_2(t)\leq y_{2,max},\quad Z_c r_2(t) \leq r_{2,max}\,.
\label{eq:constraints}
\end{equation}
where $V_c$ and $Z_c$ are matrices of appropriate size that can represent lower and upper limits. ILC seeks to update the feedforward signal $r_2(t)$ based on data gathered during previous iterations. The ILC implementation in this work is based on \cite{Schoellig2012}.

The goal is to track a desired trajectory $y^*_2(t)$ over a finite-time interval. The desired output trajectory is assumed to be feasible based on the nominal model \eqref{eq:nominal} and the constraints in \eqref{eq:constraints}; that is, there exist nominal reference, state and output trajectories $(r^*_2(t),x^*(t),y^*_2(t))$ that satisfy \eqref{eq:nominal} and \eqref{eq:constraints}. We also assume that the system stays close to the reference trajectory; hence, we only consider small deviations from the above nominal trajectories, $\tilde{r}_2(t)$, $\tilde{x}(t)$ and $\tilde{y}_2(t)$, respectively. The system is linearized about the nominal trajectory to obtain a time-varying, linear state-space model, which approximates the system dynamics along the reference trajectory. This system is then discretized and written as: 
\begin{equation}
\tilde{x}(k+1) = A(k)\tilde{x}(k)+B(k)\tilde{r}_2(k)\,,
\label{eq:discretized}
\end{equation}
where $k\in\{0,1,\hdots,N-1\}$, $N<\infty$, represents the discrete-time index. 

Using the lifted representation introduced in Section~\ref{sec:problem}, we define $\bar{\bf{y}}_{2,j}=(\tilde{y}_2(1),\hdots,\tilde{y}_2(N))\in\mathbb{R}^N$ and $\bar{\bf{r}}_{2,j}=(\tilde{r}_2(0),\hdots,\tilde{r}_2(N-1))\in\mathbb{R}^N$ and write the extended system as:
\begin{equation}
\bar{\bf{y}}_{2,j}={\bf{F}}_{\text{ILC}}\bar{\bf{r}}_{2,j}+{\bf{d}}_\infty\,,
\label{eq:lifted}
\end{equation}
where the subscript $j$ represents the iteration number, ${\bf{F}}_{\text{ILC}}$ is a constant matrix derived from the discretized model \eqref{eq:discretized} as described in \cite{Schoellig2012} and ${\bf{d}}_\infty$ represents a repetitive disturbance that is initially unknown, but is identified during the learning process. The constraints of the system can be written in the lifted representation accordingly: 
\[
\bff{V}{c}\bbff{y}{2,j}\leq \bbff{y}{2,max},\quad \bff{Z}{c} \bbff{r}{2,j} \leq \bbff{r}{2,max}\,,
\]
where $\bff{V}{c}$ and $\bff{Z}{c}$ are matrices of appropriate size.

We follow the learning approach presented in \cite{Mueller2012} and \cite{Schoellig2012} for a single system. An iteration-domain Kalman filter for the system \eqref{eq:lifted} is used to compute the estimate $\widehat{\bf{d}}_j$ based on measurements from iterations $1,\hdots,j$. The disturbance estimate is obtained from a Kalman filter based on the following model:
\begin{equation}
\begin{array}{c}
{\bf{d}}_{j+1}={\bf{d}}_j+\omega_j \\
\bar{\bf{y}}_{2,j} = {\bf{F}}_{\text{ILC}}\bar{\bf{r}}_{2.j}+{\bf{d}}_j+\mu_j\,,
\end{array}
\label{eq:systemmodel}
\end{equation}
where $\omega_j\sim \mathcal{N}(0,{\bf{E}}_j)$ and $\mu_j\sim \mathcal{N}(0,{\bf{H}}_j)$. The covariances ${\bf{E}}_j$ and ${\bf{H}}_j$ may be regarded as design parameters to adapt the learning rate of the algorithm. A common choice are diagonal covariances, such that ${\bf{E}}_j=\eta{\bf{I}}$ and ${\bf{H}}_j=\epsilon{\bf{I}}$, where $\eta\, ,\ \epsilon \in \mathbb{R}$ and $\bf{I}$ represents an identity matrix of appropriate size. The estimation equations are: 
\begin{equation}
\widehat{\bf{y}}_{j|j-1}={\bf{F}}_{\text{ILC}}\bar{\bf{r}}_{2,j}+\widehat{\bf{d}}_{j-1|j-1}\,,
\label{eq:outputerror}
\end{equation}
where 
\begin{equation}
\widehat{\bf{d}}_{j|j}=\widehat{\bf{d}}_{j-1|j-1}+{\bf{K}}_j(\bar{\bf{y}}_{2,j}-\widehat{\bf{y}}_{j|j-1})\,,
\end{equation}
and ${\bf{K}}_j$ is the optimal Kalman gain. 

An update step, based on the optimization of a cost function, computes the next input sequence $\bar{\bf{r}}_{2,j+1}$ that compensates for the identified disturbance $\widehat{\bf{d}}_{j|j}$ and estimated output error $\widehat{\bf{y}}_{j+1|j}$ in the following way: 
\begin{equation}
J(\bar{\bf{r}}_{2,j+1})=\min_{\bar{\bf{r}}_{2,j+1}\in\widehat{\bf{\Omega}}_{j+1}}\left[ \widehat{\Phi}_{j+1}\triangleq \frac{1}{2} \left\{ \widehat{\bf{y}}_{j+1|j}^T{\bf{Q}}\widehat{\bf{y}}_{j+1|j} + 
\bar{\bf{r}}_{2,j+1}^T{\bf{W}}\bar{\bf{r}}_{2,j+1} \right\} \right]\,,
\label{eq:costfunction}
\end{equation}
subject to 
\begin{equation}
\bff{V}{c}\hbff{y}{j+1|j}\leq \hbff{y}{max},\quad \bff{Z}{c} \bbff{r}{2,j} \leq \bbff{r}{2,max}\,,
\label{eq:liftedconst}
\end{equation} 
where $\bff{V}{c}$ and $\bff{Z}{c}$ are matrices of appropriate size and $\widehat{\bf{y}}_{j+1|j}$ is defined in \eqref{eq:outputerror}. The set $\widehat{\bf{\Omega}}_{j+1}$ is a convex set defined by the constraints in \eqref{eq:liftedconst}. The constant matrix ${\bf{Q}}$ is symmetric positive definite, and the constant matrix $\bff{W}{}$ is symmetric positive semidefinite and both weight different components of the cost function. The cost function tries to minimize the tracking error of the system (weighted by ${\bf{Q}}$) and a function of the control effort (weighted by ${\bf{W}}$). The resulting convex optimization problem can be solved very efficiently with state-of-the-art optimization libraries. A common approach is to define the weighting matrix as $\bff{W}{}=w\bff{I}{}$, where $w\in \mathbb{R}$ and $\bff{I}{}$ represents an identity matrix of appropriate size. In Section~\ref{sec:problem} we defined the cost function \eqref{eq:costfunctione} which tried to minimize the error $\bf{e}$. In equation~\eqref{eq:costfunction} we specify a cost function that tries to minimize the estimate $\hbff{y}{j+1,j}$ of the error $\bf{e}$ and further ensures that a smooth and executable reference input is obtained as a result of the optimization process.

To prove the asymptotic zeroing of the tracking error under  the constrained, optimization-based ILC, we make the following assumptions: 
\begin{assumption}[Rank of ${\bf{F}}_{\text{ILC}}$]
\label{ass:Frank}
The matrix ${\bf{F}}_{\text{ILC}}$ has full row-rank.
\end{assumption}
If ${\bf{F}}_{\text{ILC}}$ does not have full row-rank, a projection operator onto the image space of ${\bf{Q}}^{\frac{1}{2}}\bff{F}{\text{ILC}}$ must be introduced in order to prove convergence of the controllable part of the system \cite{Lee2000}. 
\begin{assumption}[Input constraints]
\label{ass:inputconstraint}
Given the input constraints in \eqref{eq:liftedconst}, reference trajectory ${\bf{y}}_2^*=({y}^*_2(1),\hdots,{y}^*_2(N))\in\mathbb{R}^N$, and the actual steady-state disturbance ${\bf{d}}_\infty$, the zeroing of the error is possible with an input $\bbff{r}{2,\infty}$ in the feasible set. We further assume that the active equality constraints are defined such that $[\bff{V}{c,act}\bff{F}{\text{ILC}}\ \ \bff{Z}{c,act}]^T$ is full rank.
\end{assumption}
In other words, there exists $\bar{\bf{r}}_{2,\infty}$ such that ${\bf{F}}_{\text{ILC}}\bar{\bf{r}}_{2,\infty}+{\bf{d}}_\infty=0$. In addition, \eqref{eq:liftedconst} holds.

\begin{remark}
Under Assumptions \ref{ass:Frank} and \ref{ass:inputconstraint}, system \eqref{eq:lifted} converges to the global minimum under the Kalman-filter based, constrained optimization ILC with \eqref{eq:costfunction} and \eqref{eq:liftedconst}.
\label{rem:convergence}
\end{remark}

\rev{A discussion of Remark~\ref{rem:convergence} is found in Appendix~\ref{app:remconvergence}.}
\subsubsection{Multi-Input Multi-Output Implementation:}
\label{sssec:MIMOILC}
The SISO architecture derived so far can be extended to a multi-input multi-output (MIMO) implementation. We make use of the assumption that states are decoupled (after applying an appropriate feedback linearization). Hence, the control input is a vector $\bff{r}{2,j}\in\mathbb{R}^{n_u}$ and the output is a vector $\bff{y}{2,j}\in\mathbb{R}^{n_y}$. Moreover, the matrices $A$ and $B$ are implemented as: 
\[
\begin{array}{lr}
A = \text{diag } (A_1,\hdots,A_n)\,,
& B = \text{diag }(B_1,\hdots,B_n)\,. 
\end{array}
\]

In the lifted representation we define the input and the output as 
\[
\begin{split}
\bbff{r}{2,j}=({r}_{2,1}(0), \hdots, {r}_{2,n_u}(0),\hdots,{r}_{2,1}(N-1), \hdots, {r}_{2,n_u}(N-1) ) \\
\bbff{y}{2,j}=({y}_{2,1}(1), \hdots, {y}_{2,n_y}(1),\hdots,{y}_{2,1}(N), \hdots, {y}_{2,n_y}(N) )\,, 
\end{split}
\]
respectively. We modify $\bff{F}{\text{ILC}}$ accordingly. Finally, we redefine the weighing matrices in the cost function as: 
\[
\begin{array}{rcl}
\bff{Q}{} & = & \text{diag}(\text{diag}(q_1,\hdots,q_{n_y}),\hdots,\text{diag}(q_1,\hdots,q_{n_y}))\\
\bff{R}{} & = & \text{diag}(\text{diag}(r_1,\hdots,r_{n_u}),\hdots,\text{diag}(r_1,\hdots,r_{n_u}))\\
\bff{S}{} & = & \text{diag}(\text{diag}(s_1,\hdots,s_{n_u}),\hdots,\text{diag}(s_1,\hdots,s_{n_u}))\,.
\end{array}
\]

\rev{
\subsection{Transfer Learning}
\label{ssec:transfer}
The purpose of transfer learning is to exchange learned trajectories between dynamically different systems and achieve a performance comparable to the one obtained from learning in the training system. The $\lone$ adaptive controller makes two systems behave in a repeatable predefined way, even under unknown and changing disturbances. Therefore, learned trajectories can usually be exchanged without any modification when the underlying reference model \eqref{eq:referencesystem} is the same for both systems. The learned input $\bar{\bf{r}}_{2.j}$ on the training system can be transfered without any modifications to the new system. In order to allow the new system to continue learning after the initial transfer we need to provide the ILC with an initial estimate of the repetitive disturbance $\bf{d}_j$ which is also obtained from the training system without any modifications. Equation \eqref{eq:systemmodel} to \eqref{eq:liftedconst} compute the next input sequence $\bar{\bf{r}}_{2.j+1}$ such that the system continues learning.}

%% file: Experiments.tex
This section shows the experimental results of the proposed framework combining $\lone$ adaptive control and ILC ($\lone$-ILC) applied to quadrotors for high-accuracy trajectory tracking. We compare the performance of the proposed framework to the performance of two baseline controllers: a PD (proportional-derivative) controller combined with ILC (PD-ILC) and a PID (proportional-integral-derivative) controller combined with ILC (PID-ILC). We consider four scenarios 
\begin{enumerate*}[font=\itshape] 
\item learning under unknown and changing disturbances, 
\item transfer learning between dynamically different systems, 
\item transfer learning from simulation to real-world experiments, and 
\item initializing the robot learning with a reference input generated  based on the $\lone$ adaptive controller reference model. 
\end{enumerate*}

Section \ref{sec:experimental_setup} describes the experimental setup, introduces the two quadrotors used in this study, and compares their dynamical behavior under $\lone$, PD, and PID control.
Section \ref{sec:experiment1} discusses the tracking performance under changing conditions.
Section \ref{sec:experiment2} focuses on the transferability of learned trajectories between dynamically different quadrotors. The transferability from simulation to real world is assessed in Section \ref{sec:experiment3}, and Section \ref{sec:experiment4} discusses the ability to compute the initial reference input assuming that the system behaves as the $\lone$ reference model.

\subsection{Experimental Setup}
\label{sec:experimental_setup}

\begin{figure}[t]
\center
\includegraphics[width=.6\textwidth]{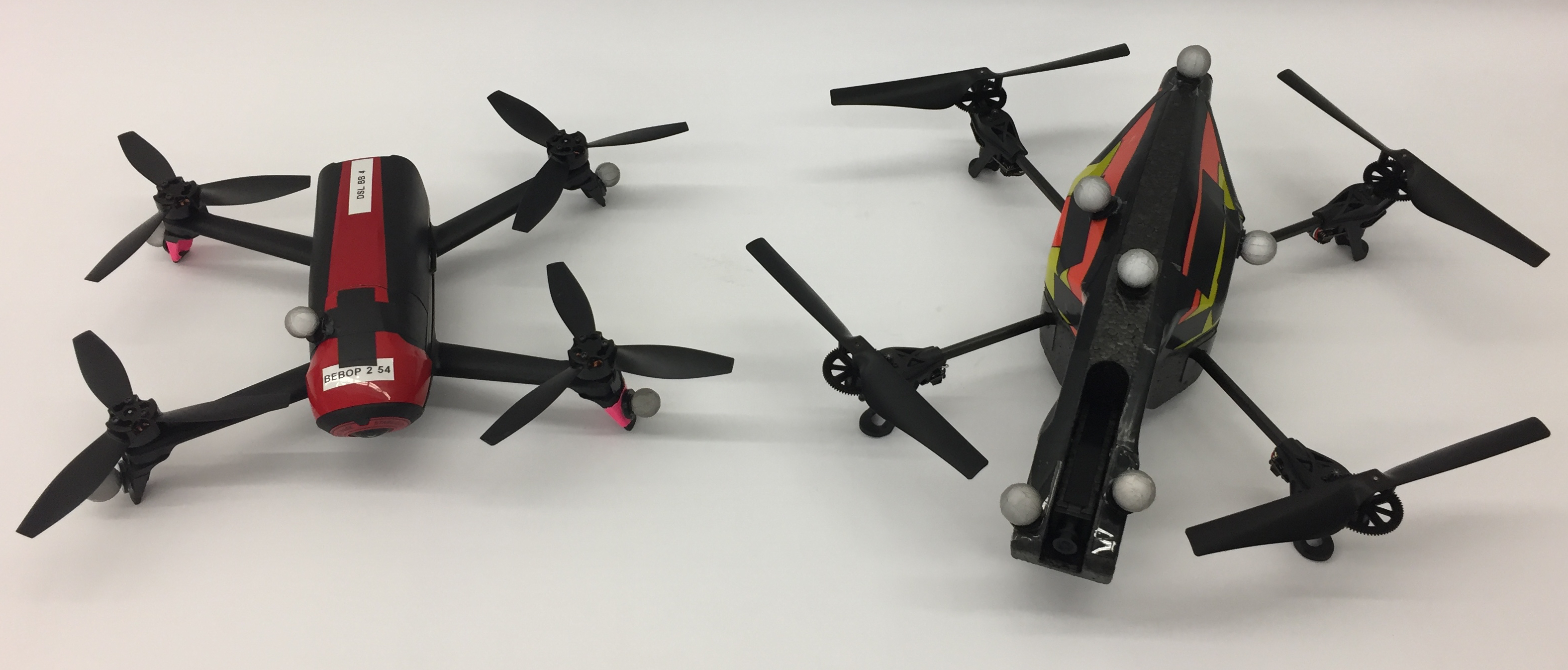}
\caption{Vehicles used in the experiments. On the left the Bebop 2, on the right the AR.Drone 2.0.}
\label{fig:drones}
\end{figure}

The vehicles used in the experiments are the Parrot AR.Drone 2.0 and the Parrot Bebop 2 (see Fig.~\ref{fig:drones}). The signals $r_1(t)$, $r_2(t)$, $y_1(t)$, and $y_2(t)$ in Fig.~\ref{fig:blockdiagram} are here the desired translational velocity, desired position, quadrotor translational velocity and quadrotor position, respectively. We implement a MIMO extended $\lone$ adaptive controller for position control as described in Section~\ref{sssec:MIMO}, where we assume that the $\mathsf{x}$, $\mathsf{y}$, and $\mathsf{z}$ directions are decoupled. A central overhead motion capture camera system provides, velocity, roll-pitch-yaw Euler angles and rotational velocity measurements. The output of the extended $\lone$ adaptive controller ${\bf{u}}(t) = (u_\mathsf{x}(t),u_\mathsf{y}(t),u_\mathsf{z}(t))$, commanded $\mathsf{x}$ and $\mathsf{y}$ translational acceleration and commanded $\mathsf{z}$ velocity, respectively, is specified in the global coordinate frame. However, the interface to the real quadrotor ('Plant' in Fig.~\ref{fig:blockdiagram}) requires commanded roll ($\phi_{des}$), pitch ($\theta_{des}$), vertical velocity ($\dot{z}_{des}$), and rotational velocity around the $\mathsf{z}$ axis ($\omega_\mathsf{z}$) (see \cite{Powers2015}). Therefore, the signal ${\bf{u}}(t)$ is transformed through the following nonlinear transformation 
\begin{align*}
\phi_{des} &= -\arcsin \left( -u_\mathsf{x} \sin(\psi) + u_\mathsf{y} \cos(\psi) \right) \\
\theta_{des} &= \arcsin \left( u_\mathsf{x} \cos(\psi) + u_\mathsf{y} \sin(\psi) \right) \\
\dot{z}_{des} &= u_\mathsf{z}\,,
\end{align*}
where $\psi$ is the current yaw angle. During the experiment, the desired yaw angle ($u_\psi$) is set to zero and controlled through a simple proportional controller $u_{\omega_\mathsf{z}} = k_\psi(u_\psi-\psi)$, where $k_\psi$ is the control gain.

We implement three different position controllers for comparison. For the extended $\lone$ adaptive controller, the controller parameters for the adaption rate $\Gamma$, reference model eigenvalues $m_\mathsf{x}$, $m_\mathsf{y}$, and $m_\mathsf{z}$, respectively and gain matrix $K$ are given in Table ~\ref{tbl:parameters}. We choose a first-order low-pass filter $C(s) = \text{diag}(\frac{\omega_\mathsf{x}}{s+\omega_\mathsf{x}},\frac{\omega_\mathsf{y}}{s+\omega_\mathsf{y}},\frac{\omega_\mathsf{z}}{s+\omega_\mathsf{z}})$. The low-pass filter is tuned for each quadrotor separately; the parameters are given in Table ~\ref{tbl:parameters2}.

\begin{table}[h]
\renewcommand*{\arraystretch}{1}
\begin{minipage}[t]{0.49\linewidth}
\centering
\begin{tabular}{ll}
\toprule
Parameter          		& Value  \\
\midrule
$\Gamma$           		& $5000$ \\
$m_\mathsf{x}$					& -1.1	 \\
$m_\mathsf{y}$					& -1.1	 \\
$m_\mathsf{z}$					& -1.75	 \\
$k_\mathsf{x}$					& 0.4	 \\
$k_\mathsf{y}$					& 0.4	 \\
$k_\mathsf{z}$					& 0.4	 \\
\bottomrule
\end{tabular}
\caption{Parameters used in the extended $\lone$ adaptive controller.}
\label{tbl:parameters}
\end{minipage}
\hfill
\begin{minipage}[t]{0.49\linewidth}
\centering
\begin{tabular}{lll}
\toprule
Parameter          		& AR.Drone 2.0 & Bebop 2 \\
\midrule
$\omega_\mathsf{x}$ & 3.5  & 23	\\
$\omega_\mathsf{y}$ & 3.5  & 23	\\
$\omega_\mathsf{z}$ & 3.5  & 3.8	\\
\bottomrule
\end{tabular}
\caption{Drone-dependent $\lone$ adaptive controller parameters for the low-pass filter $C(s) = \text{diag}(\frac{\omega_\mathsf{x}}{s+\omega_\mathsf{x}},\frac{\omega_\mathsf{y}}{s+\omega_\mathsf{y}},\frac{\omega_\mathsf{z}}{s+\omega_\mathsf{z}})$.}
\label{tbl:parameters2}
\end{minipage}
\end{table}

We compare the performance of the proposed $\lone$-ILC approach with that of PD-ILC and PID-ILC. The PD controller is given by:
\begin{equation}
u_i(t) = \frac{2\zeta}{\tau_i}(\dot r_{2,i}(t) - y_{1,i}(t)) + \frac{1}{\tau^2_i}(r_{2,i}(t) - y_{2,i}(t))\,, \quad \text{for} \; i = \mathsf{x},\,\mathsf{y},\,\mathsf{z}\,, 
\end{equation}
where $\tau_i$ and $\zeta$ are the time constant and damping coefficient, respectively.
The PID controller is given by:
\begin{equation}
u_i(t) = \alpha(\dot r_{2,i}(t) - y_{1,i}(t)) + \beta(r_{2,i}(t) - y_{2,i}(t)) + \gamma\int(r_{2,i}(t) - y_{2,i}(t))dt\,, \quad \text{for} \; i = \mathsf{x},\,\mathsf{y},\,\mathsf{z}\,, 
\end{equation}
where $\alpha$, $\beta$, and $\gamma$ are the controller gains which could be defined as in \cite{schoellig2010}: $\alpha = \tau_i(1+2\zeta)$, $\beta = \tau_i^2(1+2\zeta)$, and $\gamma = \tau_i^3$.

In this application, constraints are imposed on the input acceleration as it is intimately related to the physical capabilities of the actuators of the system and are expressed through the following mathematical inequality:   
\begin{equation}
\ddot{\bf{r}}^{\text{low}}\leq \ddot{\bar{\mathbf{r}}}_{2,j+1} \leq \ddot{\bf{r}}^{\text{hi}}\,,
\label{eq:accconstraint}
\end{equation}
where the sequence $\ddot{\bar{\mathbf{r}}}_{2,j+1}$ represents the discrete approximation of the second derivative of the input reference. The above constraint can be rearranged as linear inequality with respect to $\bar{\bf{r}}_{2,j+1}$. Under the assumption that $\ddot{\bar{r}}_{2,j+1}(N) = \ddot{\bar{r}}_{2,j+1}(N-1)$, $\ddot{\bar{\mathbf{r}}}_{2,j+1}$ can be written as: 
\begin{equation}
\ddot{\bar{\mathbf{r}}}_{2,j+1} = 
\begin{bmatrix}
({\bar{r}}_{2,j+1}(2)-2{\bar{r}}_{2,j+1}(1)+{\bar{r}}_{2,j+1}(0))/{(\Delta t)^2} \\
({\bar{r}}_{2,j+1}(3)-2{\bar{r}}_{2,j+1}(2)+{\bar{r}}_{2,j+1}(1))/{(\Delta t)^2}  \\ 
\vdots \\
({\bar{r}}_{2,j+1}(N)-2{\bar{r}}_{2,j+1}(N-1)+{\bar{r}}_{2,j+1}(N-2))/{(\Delta t)^2} 
\end{bmatrix}
= {\bf{D}}\ddot{\bar{\mathbf{r}}}_{2,j+1}\,,
\label{eq:discreteacc}
\end{equation}
where 
\begin{equation}
{\bf{D}}=
\begin{bmatrix}
1 /{(\Delta t)^2}& -2 /{(\Delta t)^2}& 1 /{(\Delta t)^2}& 0 & \hdots & 0 \\
0 & 1/{(\Delta t)^2} & -2/{(\Delta t)^2} & 1/{(\Delta t)^2} & \ddots & 0 \\
\vdots &  & \ddots & \ddots & \ddots & 0 \\
0 & 0 & \hdots & 1 /{(\Delta t)^2}& -2/{(\Delta t)^2} & 1/{(\Delta t)^2}
\end{bmatrix}\,.
\label{eq:discretedderivative}
\end{equation}
and $\Delta t$ is the time interval between discrete samples. 

In this implementation we define a cost function that minimizes the estimated error $\hbff{y}{j+1|j}$ while achieving a smooth input with the minimum control effort $\hbff{r}{2,j+1}$. Hence, we include the estimated error, the control effort and input accelerations in the cost function of this implementation:  
\begin{equation}
J(\bar{\bf{r}}_{2,j+1})=\min_{\bar{\bf{r}}_{2,j+1}\in\widehat{\bf{\Omega}}_{j+1}}\left[ \widehat{\Phi}_{j+1}\triangleq \frac{1}{2} \left\{ \widehat{\bf{y}}_{j+1|j}^T{\bf{Q}}\widehat{\bf{y}}_{j+1|j} + 
\bar{\bf{r}}_{2,j+1}^T{\bf{W}}\bar{\bf{r}}_{2,j+1} \right\} \right]\,,
\label{eq:costfunctionspecific}
\end{equation}
subject to \eqref{eq:accconstraint}, where $\bff{Q}{}=\bff{I}{}$. Moreover, we define $\bff{W}{}={\bf{R}}+{\bf{D}}^T{\bf{S}}{\bf{D}}$ to penalize control effort (weighted by ${\bf{R}}=r\bff{I}{}$ with $r=0.001$) and the acceleration of the reference signal (weighted by ${\bf{S}}=s\bff{I}{}$ with $s=0.0025$). We use the IBM CPLEX optimizer to solve the above optimization problem. Using this definition, it can be shown that ${\bf{W}}$ is symmetric positive definite. 

If the constraints \eqref{eq:accconstraint} are inactive, according to Remark~\ref{rem:convergence}, system \eqref{eq:lifted} converges to the global minimum under the Kalman-filter based, constrained optimization ILC approach. However, if constraints are active, they are included in the Lagrangian in the following way:
 \[
 \begin{array}{rcl}
  \mathcal{L}(\bar{\bf{r}}_{2,j}, {\bm{\lambda}}_1, {\bm{\lambda}}_2) & = &   \frac{1}{2} \left \{ ({\bf{F}}_{\text{ILC}} \bar{\bf{r}}_{2,j+1} + \widehat{\bf{d}}_{j|j} )^T{\bf{Q}}({\bf{F}}_{\text{ILC}} \bar{\bf{r}}_{2,j+1} + \widehat{\bf{d}}_{j|j} ) + \bar{\bf{r}}_{2,j+1}^T{\bf{W}}\bar{\bf{r}}_{2,j+1}\right\} \\
  & & - \sum_{l\in M}{\lambda}_{1,l} ( {\bf{D}}_l \bar{\bf{r}}_{2,j+1} - {\ddot{{r}}}^{\text{hi}} ) - \sum_{l\in P} {\lambda}_{2,l} ( {\bf{D}}_l \bar{\bf{r}}_{2,j+1} - {\ddot{{r}}}^{\text{low}})  \\
 \end{array}  
 \]
where ${\bf{D}}_l$ is the $l^{th}$ row of \eqref{eq:discretedderivative} that corresponds to an active constraint, ${{\lambda}}_{1,l}$ and ${{\lambda}}_{2,l}$ are Lagrange multipliers for the set $M$ of maximum acceleration and the set $P$ of minimum acceleration active constraints. Assumption~\ref{ass:inputconstraint} holds because at any given point in the trajectory only one set of constraints, either minimum or maximum, can be active. Hence, $M\cap P=\{0\}$ and $\bff{Z}{c,act}$, the matrix whose rows are the rows of ${\bf{D}}$ that correspond to the active constraints, is full rank. We can conclude then that $\bbff{r}{2,j+1}$ is the unique global solution to the minimization problem. 
 
\begin{figure}[t]
\includegraphics[width=\textwidth]{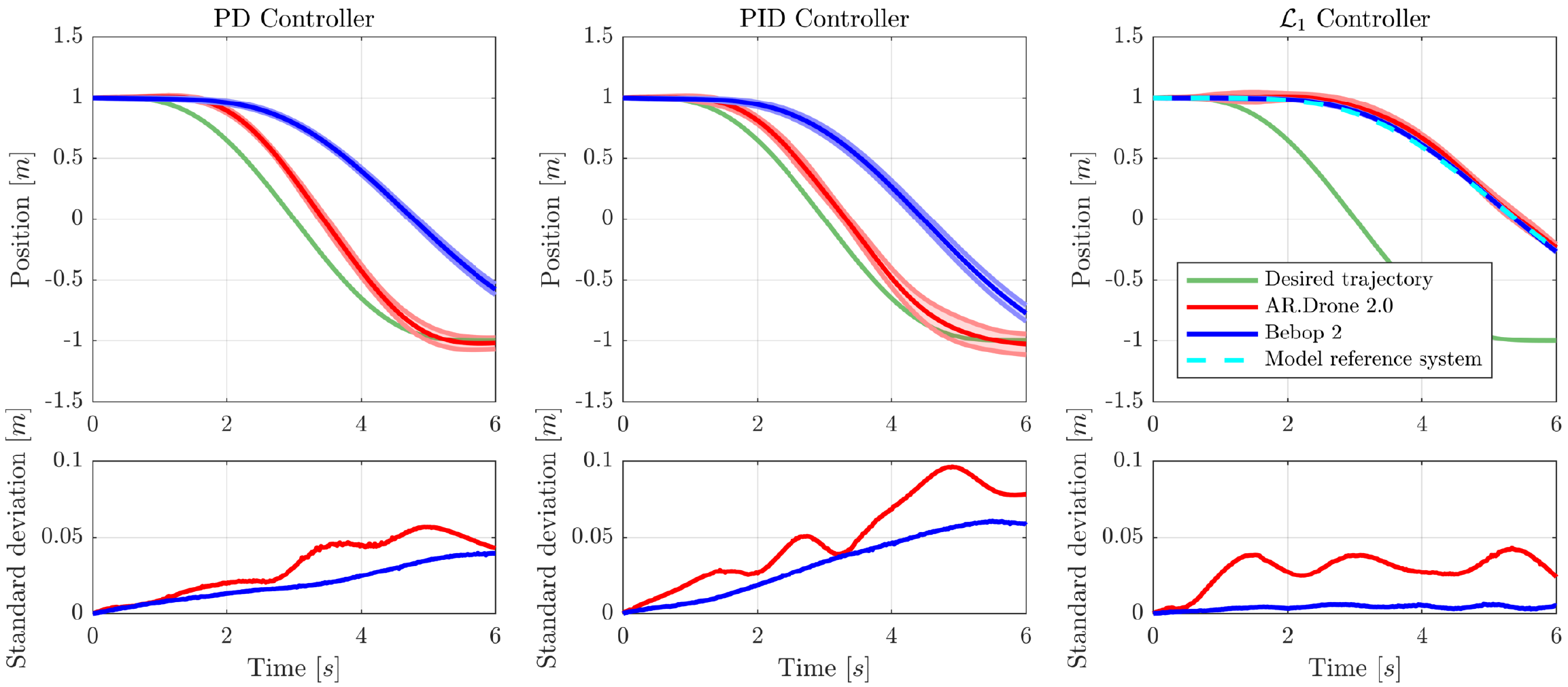}
\caption{Time response in $\mathsf{x}$ direction for the AR.Drone 2.0 and Bebop 2 for a given reference trajectory using three different controllers: (left) PD, (middle) PID, (right) $\lone$. The response of the model reference system of the $\lone$ controller is depicted in the plot of the $\lone$ controller. The line in the top figures denotes the mean over five repetitions, and the envelope denotes the standard deviation. The bottom figures shows the standard deviation over time. It can be seen that with the use of the PD and PID controller the drones have different dynamic behavior, whereas the $\lone$ adaptive controller forces the systems to behave as the model reference system.}
\label{fig:repeatability}
\end{figure}

To show that the two quadrotors have different dynamical behavior, we use the same controller gains for the PD and PID controller for both quadrotors. Each of the two quadrotors is tasked to track a three-dimensional straight line reference trajectory using the PD, PID, and $\lone$ controller. Fig.~\ref{fig:repeatability} compares the time response in $\mathsf{x}$ direction of the two quadrotors for each controller. The dynamical difference between the AR.Drone 2.0 and Bebop 2 are significant for both the PD and PID controller, while using the extended $\lone$ adaptive controller, both drones behave similarly and close to the $\lone$ model reference system. This confirms that the $\lone$ adaptive controller framework implemented as an underlying controller enforces the same dynamic behavior on dynamically different systems. It is also interesting to observe that for repeated experiments, the standard deviation (Fig.~\ref{fig:repeatability}, lower row) over time stays constant for the $\lone$ controller and increases with time for the PD and PID controller. This shows that the $\lone$ controller renders a more repeatable system overall.

To quantify the performance of the ILC, an average position error along the trajectory is defined as: 
\begin{equation}
e = \frac{\sum\limits_{i=1}^N \sqrt{(r_{2,\mathsf{x}}(i)-y_{2,\mathsf{x}}(i))^2 + (r_{2,\mathsf{y}}(i)-y_{2,\mathsf{y}}(i))^2 + (r_{2,\mathsf{z}}(i)-y_{2,\mathsf{z}}(i))^2}}{N}\,.
\label{eq:ILC_error}
\end{equation}

\subsection{Learning Under Disturbance}
\label{sec:experiment1}

To asses the performance under changing conditions, an external wind is introduced as a disturbance. This wind is generated by a fan placed on the floor blowing wind in the direction perpendicular to the trajectory path of the quadrotor. 

\begin{figure}[t]
\centering
\includegraphics[width=0.49\textwidth]{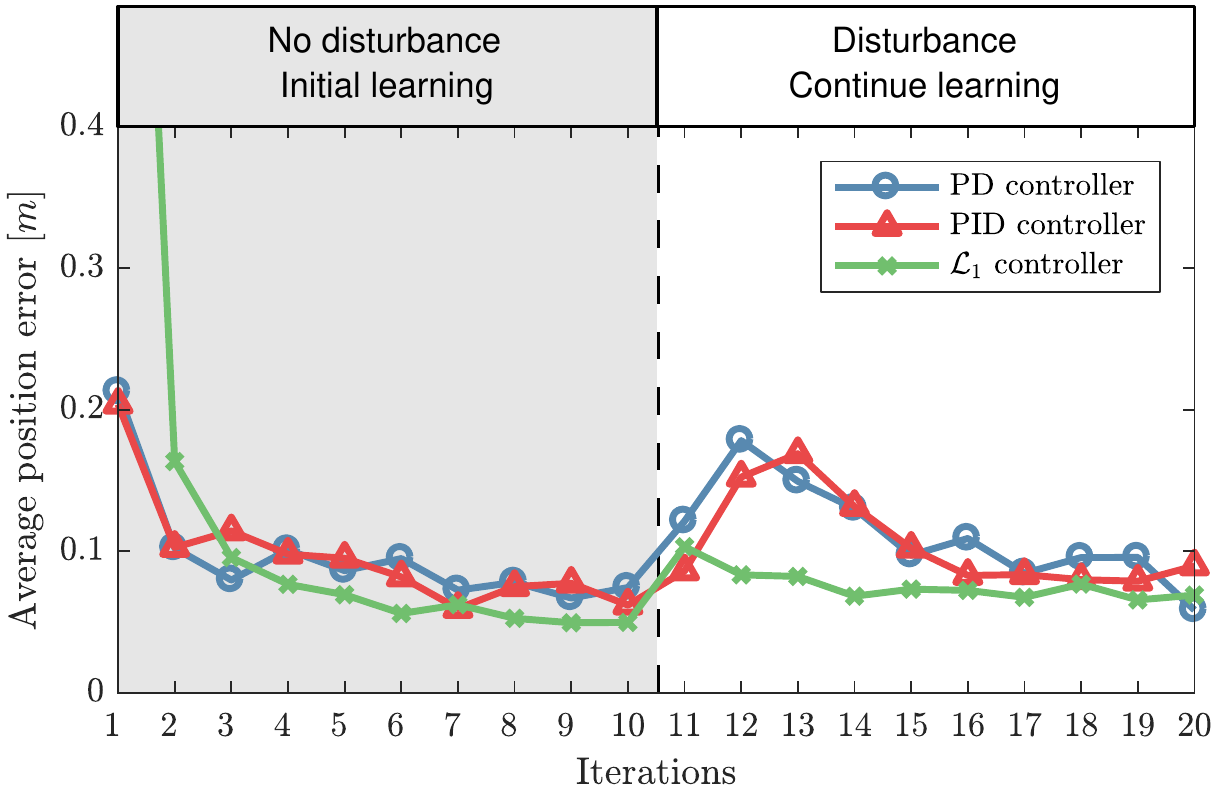}
\caption{Mean of the average position error \eqref{eq:ILC_error} over five separate learning experiments using the PD-ILC, PID-ILC and $\lone$-ILC frameworks. No disturbance is applied in iteration 1-10. After iteration 10 an external disturbance (wind) is applied and learning is continued for iteration 11-20. The initial value of the error using the $\lone$ controller is $0.99$ [$m$] in iteration 1, which is larger than the PD and PID controllers because of the relatively slow model reference system. After the wind disturbance is applied, the PD-ILC and PID-ILC frameworks must relearn and show a significantly larger error in iteration 11 than the $\lone$-ILC setup. Both the PD and PID controllers show that the error increases within the first 3 iterations after the disturbance is applied, caused by the dynamical changes for which the ILC is not tuned, the $\lone$ controller compensates for these dynamical changes and converges quickly.}
\label{fig:wind_dist_mean}
\end{figure}

In this experiment, the AR.Drone 2.0 learns to track a desired trajectory (same diagonal trajectory as in Section~\ref{sec:experimental_setup}) using each of the three frameworks: PD-ILC, PID-ILC and $\lone$-ILC. This experiment is repeated five times and the mean of the tracking error as defined in \eqref{eq:ILC_error} for this initial learning process (iteration 1-10) is depicted in Fig.~\ref{fig:wind_dist_mean}, the average standard deviation (average over iteration 1-10) during this initial learning process is given in Table~\ref{tbl:wind_dist_std}. The proposed $\lone$-ILC shows a higher position error during the first iteration, which is expected since the model reference system is slow (see Fig.~\ref{fig:repeatability}). From iteration 4, the $\lone$-ILC shows lower errors consistently. It has also the highest repeatability (i.e. lowest standard deviation over different learning experiments), see Table~\ref{tbl:wind_dist_std}. \rev{There may be PID gains that improve the performance of the PID over the PD controller in Fig.~\ref{fig:wind_dist_mean}; however, we don't expect fundamental differences in the results.}

\begin{figure}[t]
\centering
\includegraphics[width=\textwidth]{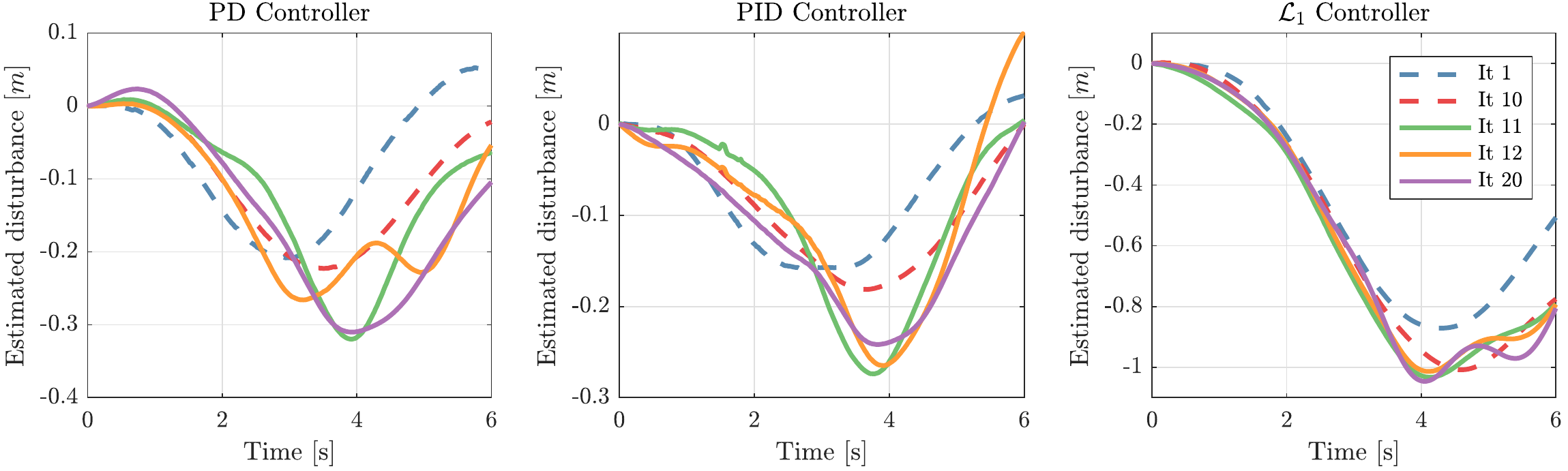}
\caption{Disturbance estimate for $\mathsf{y}$ direction obtained from the Kalman filter using the PD (left), PID (middle), and $\lone$ (right) controller. In iteration 1 and 10 (dashed lines)  no external disturbance is applied and in iteration 11, 12, and 20 (solid lines) the external wind disturbance is applied. Note that the scale for each controller is different. For the PD-ILC and PID-ILC, we can see a significantly larger change in estimated disturbance after the external wind disturbance is applied compared to the $\lone$-ILC setup.}
\label{fig:wind_dist_kalman_dist}
\end{figure}

After this initial learning process, an external wind disturbance is applied in iteration 11-20, and the ILC continues learning. While all frameworks show an increase in error in iteration 11, the $\lone$-ILC setup exhibits only a minor increase and quickly adapts to the new conditions (within two to three iterations). The PD-ILC shows the largest increase. Because of the change in dynamics caused by the disturbance, the model of the ILC is not representing the real system anymore; therefore, the error increases in iteration 12 and 13 for the PD-ILC and PID-ILC, where the $\lone$ controller is capable of adapting to this change of dynamics. Fig.~\ref{fig:wind_dist_kalman_dist} depicts the Kalman filter estimated disturbance $\hat d_j$ for the $\mathsf{y}$ direction. It can be seen that the disturbance is overestimated in iteration 11 and 12 when using the PD and PID controller, causing the error to increase in the next iteration. When using the $\lone$ controller, the estimated disturbance in the ILC component does not change much after applying the external wind disturbance since the underlying $\lone$ controller compensates for the change in dynamics. Overall, the three frameworks converge to a slightly higher average tracking error (iteration 17-20) due to the fact that the wind disturbance is partially non-repetitive (or noisy); learning is only able to compensate for systematic disturbances. Tab.~\ref{tbl:wind_dist_std} shows that the variance significantly increases when the external wind disturbance is applied within the PD-ILC framework while there is little or no increase when using the PID-ILC and $\lone$-ILC framework, respectively.


\begin{table}[h]
\centering
\begin{tabular}{lll}
\toprule
& \multicolumn{2}{c}{Average Standard Deviation [$m$]}\\
& No Disturbance & Disturbance \\
\midrule
PD-ILC & $0.0167$  & $0.0210$	\\
PID-ILC & $0.0177$  & $0.0182$	\\
$\lone$-ILC & $0.0130$ & $0.0128$	\\
\bottomrule
\end{tabular}
\caption{Average standard deviation of the tracking error over iterations. The full learning experiment (20 iterations in total) was repeated five times for each framework: PD-ILC, PID-ILC, and $\lone$-ILC.}
\label{tbl:wind_dist_std}
\end{table}

\subsection{Transfer Learning Between Dynamically Different Systems}
\label{sec:experiment2}

\begin{figure}[t]
\centering
\begin{subfigure}[b]{.49\textwidth}
  \centering
  \includegraphics[width=\linewidth]{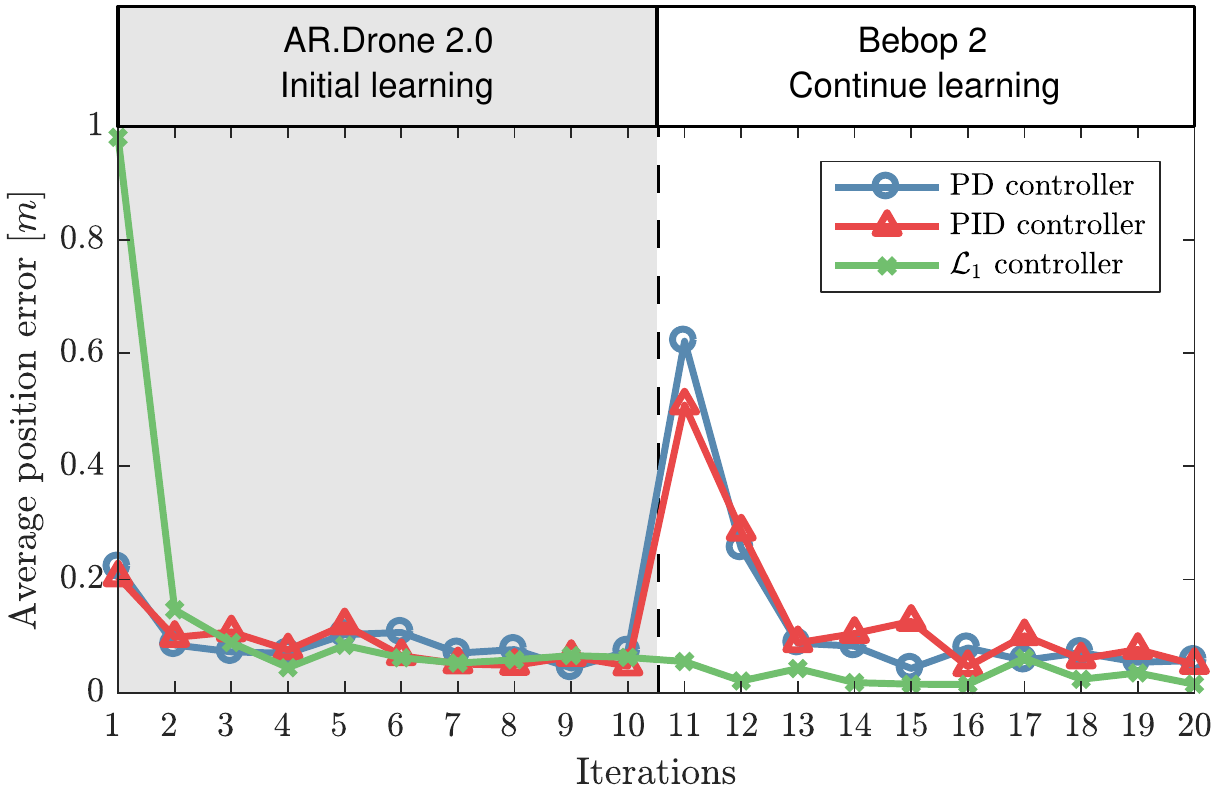}
  \caption{}\label{fig:translearn_ar_be}
\end{subfigure}
\begin{subfigure}[b]{.49\textwidth}
  \centering
  \includegraphics[width=\linewidth]{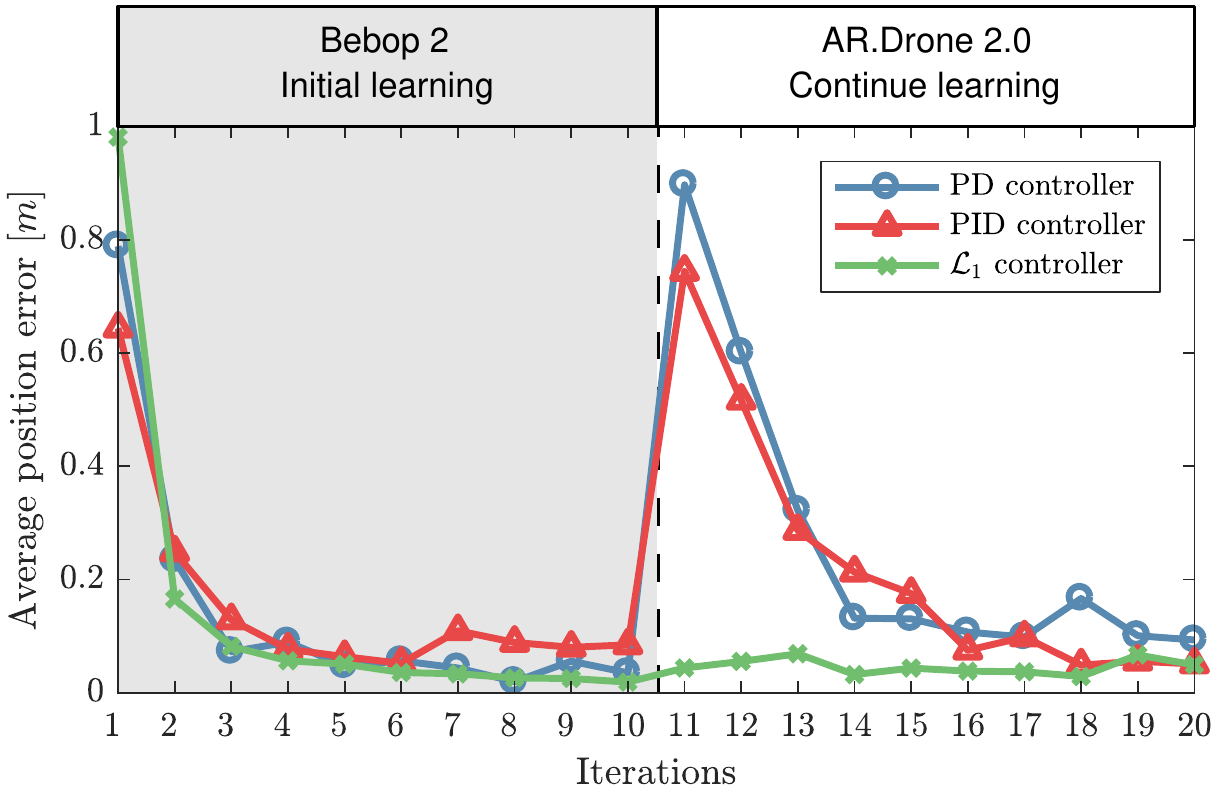}
  \caption{}
  \label{fig:translearn_be_ar}
\end{subfigure}
\caption{Learning behavior of the PD-ILC, PID-ILC and $\lone$-ILC framework when after iterations 10 the learned input and disturbance is transfered from the AR.Drone 2.0 to the Bebop 2 (a) and from Bebop 2 to AR.Drone 2.0 (b). The PD-ILC and PID-ILC approach show a significantly larger error after transfer. Note that the scale is different than in Fig.~\ref{fig:wind_dist_mean}.}
\label{fig:translearn}
\end{figure}

In this experiment, we assess the performance of transfer learning between dynamically different systems. In an initial learning phase, both the AR.Drone 2.0 and Bebop 2 quadrotors learn over ten iterations with the PD-ILC, PID-ILC, and $\lone$-ILC framework \rev{where both quadrotors use the same underlying model reference system \eqref{eq:referencesystem}}. Convergence of the error for the first ten iterations for the AR.Drone 2.0 and Bebop 2 under each control framework is shown in Fig.~\ref{fig:translearn}. After iteration 10, the learned trajectory is transferred from AR.Drone 2.0 to Bebop 2 and vice versa. Learning is continued in iteration 11-20. The increase in tracking error after transfer learning is shown in Table~\ref{tbl:performace_translearn}. The $\lone$-ILC approach shows only a marginal increase in error, while the PD-ILC and PID-ILC approach show a significant increase in error. Also can be noted that transferring the learned trajectory from a system with a low variance (Bebop 2) to a system with a larger variance (AR.Drone 2.0) increases the error. This results shows that in the $\lone$-ILC case the learned knowledge can be transferred to a dynamically different second system; the second system must be controlled by a corresponding underlying $\lone$ adaptive controller with the same reference model. More generally, this proves the potential of the $\lone$-ILC method to significantly speed up learning as one robot can learn from the other.

\begin{table}[t]
\center
\begin{tabular}{llll}
\toprule
& \multicolumn{3}{c}{Factor of Error Increase}\\
& PD-ILC & PID-ILC & $\lone$-ILC \\
\midrule
AR.Drone 2.0 to Bebop 2 & $8.492$ & $10.795$ & $0.884$	\\
Bebop 2 to AR.Drone 2.0 & $25.613$  & $8.807$ & $2.327$	\\
\bottomrule
\end{tabular}
\caption{Increase in error after transferring the learned trajectory from AR.Drone 2.0 to Bebop 2 and vice versa, for the PD-ILC, PID-ILC, and $\lone$-ILC approach. It is obvious that the $\lone$-ILC framewok handles dynamic changes significantly better (by a factor of 4 to 10).}
\label{tbl:performace_translearn}
\end{table}

\subsection{Transfer Learning from Simulation to Real System}
\label{sec:experiment3}
\begin{figure}[t]
\centering
\includegraphics[width=.49\linewidth]{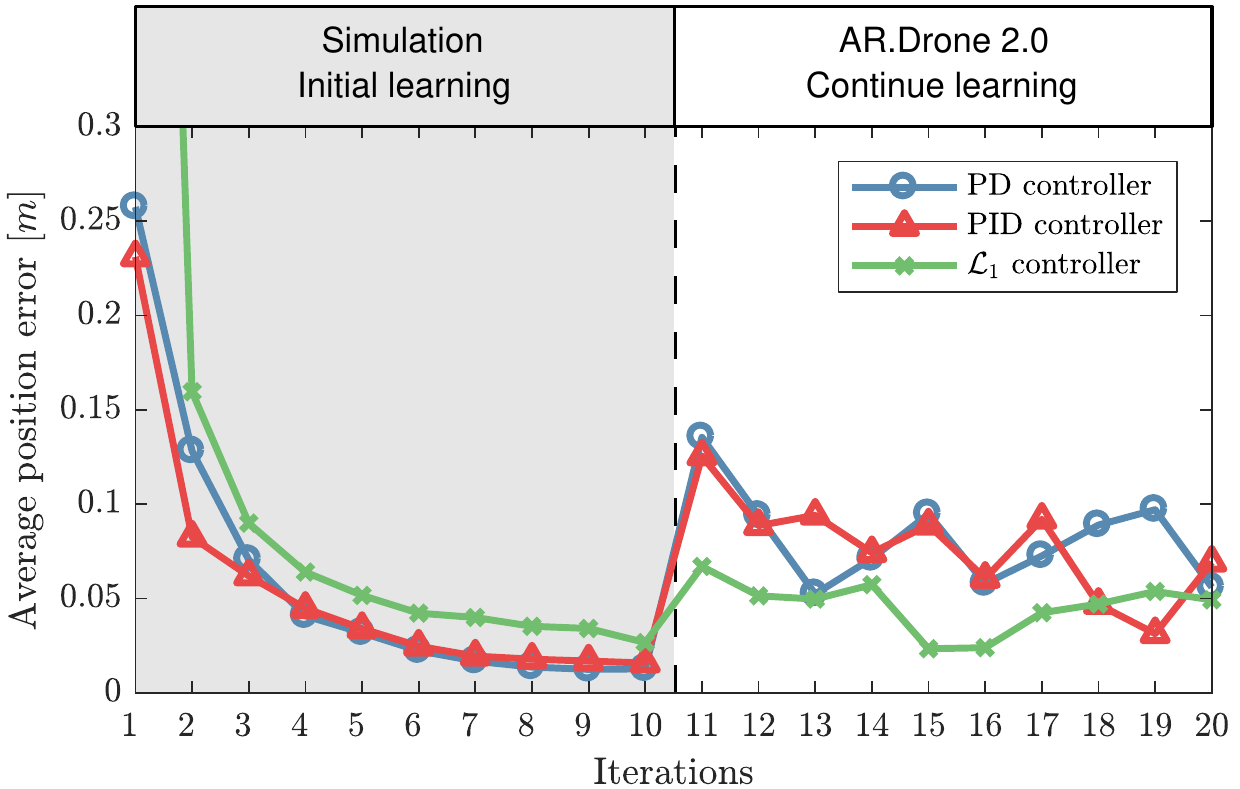}
\caption{Learning behavior of the PD-ILC, PID-ILC, and $\lone$-ILC framework in the simulator for iterations 1-10. The learned reference input and disturbance are transferred to the AR.Drone 2.0 after iteration 10, and learning continues in iteration 11-20. Note that the scale is different than in Fig.~\ref{fig:translearn_ar_be} and \ref{fig:translearn_be_ar}.}
\label{fig:translearn_sim_ar}
\end{figure}

In this experiment, we aim to assess the performance of transfer learning from simulation to real systems. The transfer performance depends on how close the simulator is to the real dynamics of the system. For this experiment, simulations have been performed in the Robot Operating System (ROS) environment using the Gazebo simulator running a simulation of the AR.Drone 2.0. Learning was performed in the simulator over 10 iterations using the PD-ILC, PID-ILC, and $\lone$-ILC approach. \rev{Here the $\lone$-ILC framework uses the same underlying model reference system \eqref{eq:referencesystem} for both the simulation and the real quadrotor.} After iteration 10, the learned trajectory is transferred to the real AR.Drone 2.0 quadrotor and ten additional learning iterations are performed. Convergence of the error is shown in Fig.~\ref{fig:translearn_sim_ar}. Again, the $\lone$-ILC framework shows the best transfer capabilities and needs no re-learning. Overall, since the simulator is very consistent (only minor sources of random noise added), it is possible to achieve very low tracking errors. With the real system, this error increases as not all disturbances are repeatable and can be compensated for. However, the AR.Drone 2.0 achieves comparable tracking errors as before, see Fig.~\ref{fig:translearn}. 

\subsection{Reference-Model Based Input to Initialize Learning}
\label{sec:experiment4}

\begin{figure}[t]
\center
\includegraphics[width=.7\linewidth]{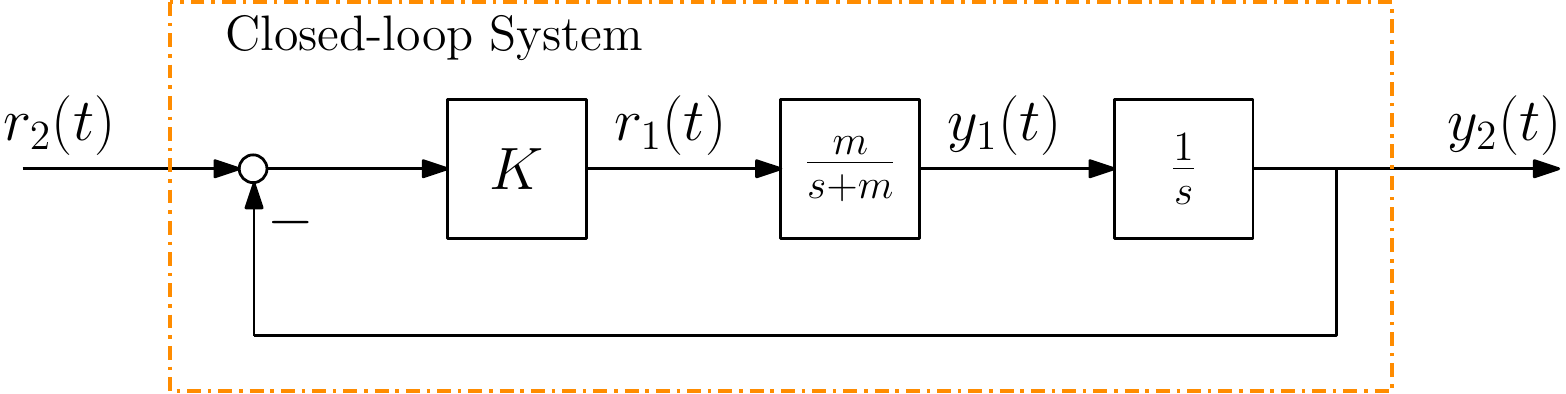}
\caption{Simplified closed-loop system (defined in Fig.~\ref{fig:blockdiagram}) that uses the fact that the system behaves like the reference model specified by the $\lone$ adaptive controller.}
\label{fig:model_reference_system}
\end{figure}

\begin{figure}[t]
\centering
\includegraphics[width=\linewidth]{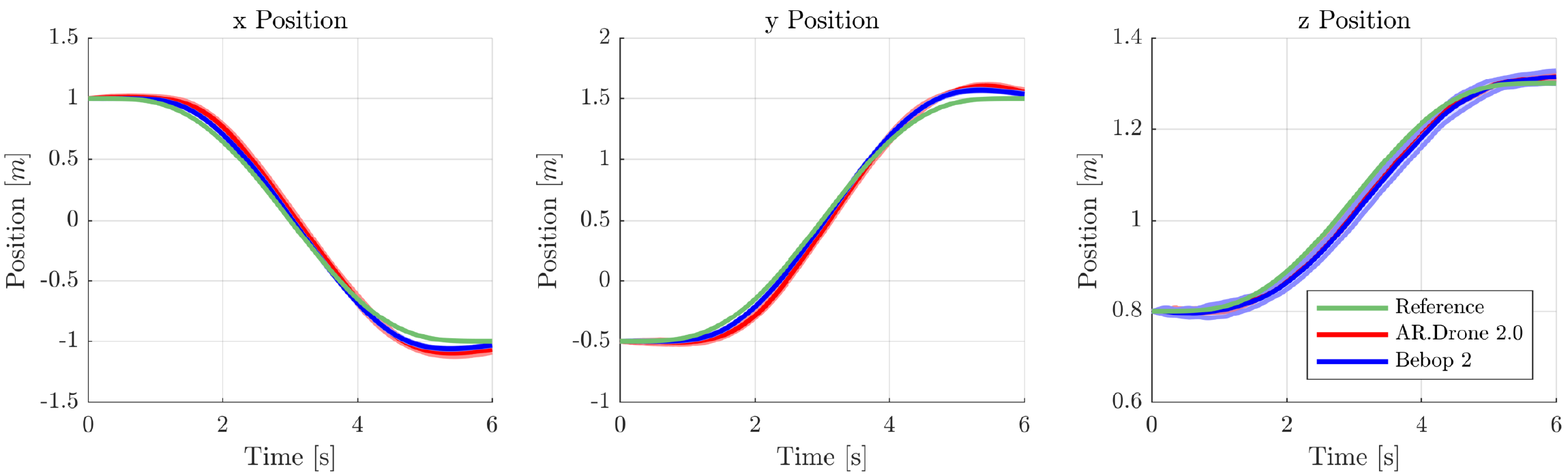}
\caption{Response over time for a three-dimensional trajectory with the input calculated based on the $\lone$ reference model applied to the AR.Drone 2.0 and Bebop 2 using the $\lone$ adaptive controller. Shown are the mean and standard deviation over 5 experiments. Both systems track the reference trajectory closely.}
\label{fig:translearn_calc_drone_repeatability}
\end{figure}

\begin{figure}[t]
\centering
\includegraphics[width=.49\linewidth]{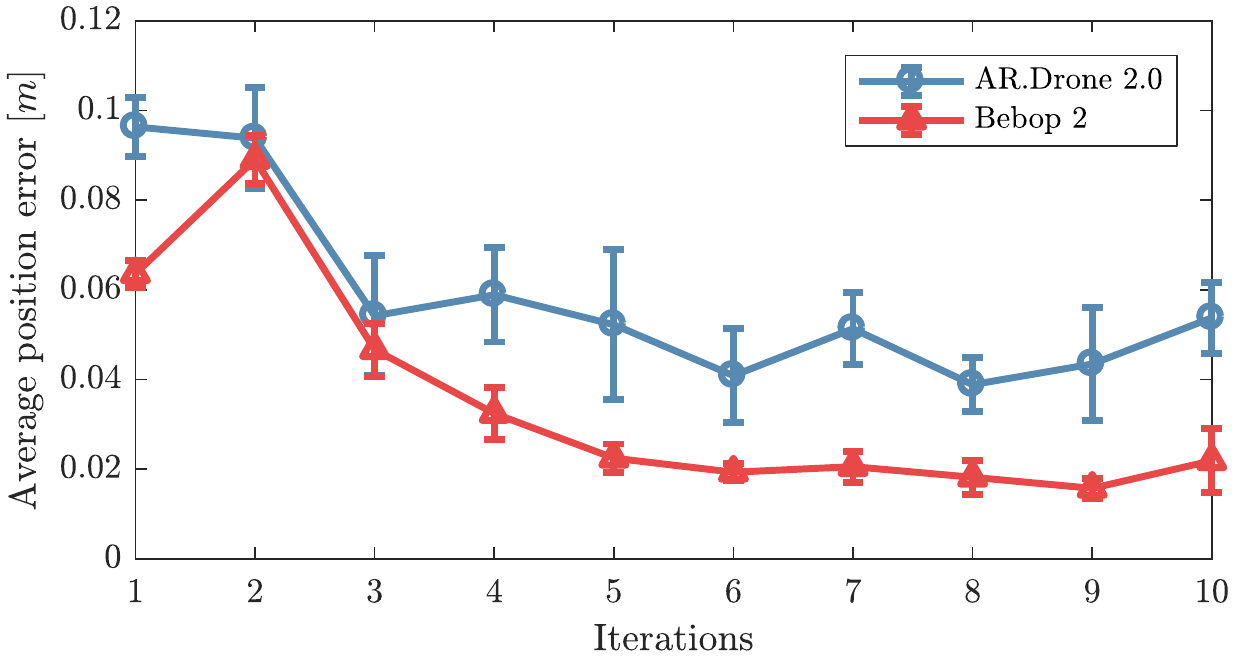}
\caption{Mean of the error for five 10-iteration sets when the $\lone$ reference model based input is used to initialize the learning for the $\lone$-ILC framework. The initial error of both systems is very low and is further reduced through learning. An average position error of 2 to 5 centimeters is very low by all standards.}
\label{fig:translearn_calc_drone}
\end{figure}

In this experiment, we want to calculate, based on the $\lone$ reference model, an initial input for the ILC. In Section~\ref{sec:experimental_setup} and Fig.~\ref{fig:repeatability} we showed that the real system behaves as the model reference system \eqref{eq:referencesystem} when using the $\lone$ adaptive controller. We use this feature to simplify the closed-loop system (as defined in Fig.~\ref{fig:blockdiagram}) and substitute the $\lone$ adaptive controller and the system with the $\lone$ reference model to obtain the block diagram shown in Fig.~\ref{fig:model_reference_system}. Using this simplified system, we obtain a state-space representation defined by: 
\[
A=
\begin{bmatrix}
0 & 1 \\
-Km & -m
\end{bmatrix}
\quad 
B=
\begin{bmatrix}
0\\Km
\end{bmatrix}\,.
\]
Furthermore, we define the initial state as $x_0$. With the above definitions, it is possible to compute an input based on \eqref{eq:referencesystem} such that the system tracks the reference exactly. Since we know the desired output $\textbf{y}_2^*$, we can compute an initial input $\textbf{r}_{2,1}$ and initial disturbance estimate $\textbf{d}_1$ using \eqref{eq:lifted}, resulting in: 

\begin{equation}
\begin{aligned}
\textbf{r}_{2,1} &= \textbf{F}_{\text{ILC}}^{-1}(\textbf{y}_2^*-\textbf{d}^0)\;, \\
\textbf{d}_1 &= \textbf{F}_{\text{ILC}}(\textbf{r}_{2,1}-\textbf{y}_2^*)\,, \\
\textbf{d}^0 &= \left[(Ax_0)^T,(A^2x_0)^T,\dots,(A^{N-1}x_0)^T\right]^T\,,
\end{aligned}
\label{eq:compute_ref}
\end{equation}
with $\textbf{d}^0$ defined as in \cite{Schoellig2012}.
Note that \eqref{eq:compute_ref} does not use the deviations from the nominal trajectories in contrast to \eqref{eq:lifted}. The calculated input $\textbf{r}_{2,1}$ is applied to both the AR.Drone~2.0 and Bebop~2 using the $\lone$ adaptive controller without learning. The response over time for a three-dimensional trajectory is shown in Fig.~\ref{fig:translearn_calc_drone_repeatability}. Since there is still a small error in the response of both drones, learning is initialized and a 10-iteration learning experiment is performed. The experiment is repeated five times, and mean and standard deviation of the error are shown in Fig.~\ref{fig:translearn_calc_drone}.

\subsection{Discussion on Input Initializing Approaches and Transfer Learning Performance}
\label{sec:discussion}

The experiments in the previous subsections demonstrate the capability of the $\lone$-ILC framework to achieve high-accuracy trajectory tracking in the first iteration of a new experiment by 
\begin{enumerate*}[font=\itshape] 
\item transferring learned experience from a dynamically different system, 
\item transferring learned experience from a simulation, and 
\item generating the first input trajectory based on the $\lone$ reference model. 
\end{enumerate*} Table~\ref{tbl:comparison} shows the initial errors and average converged errors obtained when applying methods (\textit{i}) to (\textit{iii}) described above to an AR.Drone 2.0. We compare these errors to a standard first trial where the desired output $\bf{y_2^*}$ is used as the reference input $\bf{r_2}$. We make the following observations:
\begin{itemize}
\item We begin by comparing the error achieved in the first iteration with the naive input to the input obtained by transfer learning from a different system (Section~\ref{sec:experiment2}). When the PD and PID controllers are used, the initial error of the transfer learning approach can be up to four times larger than the error of the naive input. This means that transfer learning from a dynamically different system has an adverse effect on the error for the PD and PID controllers and should not be done. In contrast, when the $\lone$ adaptive controller is used, the error in the first iteration using transfer learning from a dynamically different system is comparable to the converged error achieved after learning with the naive input. This is possible as both systems run an adaptive controller with the same, predefined model reference, which defines the system behavior. Consequently, transfer learning from a dynamically different system is highly effective for the $\lone$-ILC approach.
\item For the PD-ILC and PID-ILC approaches, using trajectories learned in simulation results in a better performance in the first iteration than transferring learning from a different real system. This is because the simulator closely resembles the behavior of the real system. In our experiments, transferring knowledge from simulation to the real system was beneficial for all frameworks compared to the initial performance with the naive input. However, partial relearning is necessary for the PD and PID case while not necessary for the $\lone$ case.
\item Using the $\lone$ adaptive controller allows us to calculate an input based on the $\lone$ reference model and to achieve an error ten times smaller than the error obtained with a naive input.
\end{itemize}
Overall, the best results are obtained with the $\lone$-ILC framework. Within this framework, using an input transferred from an initial learning process in a different system achieves the lowest tracking errors in the first iteration. For the PD-ILC and PID-ILC frameworks, it is only possible to transfer experience if the system used for initial learning closely resembles the real system. 

\begin{table}[ht]
\center
\begin{tabular}{lllllllll}
\toprule
Initial Input& \multicolumn{2}{c}{Naive} & \multicolumn{2}{c}{Transfer Learning} & \multicolumn{2}{c}{Transfer Learning} & \multicolumn{2}{c}{Calculated Input} \\
& \multicolumn{2}{c}{} & \multicolumn{2}{c}{from System} & \multicolumn{2}{c}{from Simulation} & \multicolumn{2}{c}{} \\
Iteration & $1^{\text{st}}$ & $8^{\text{th}}-10^{\text{th}}$ & $1^{\text{st}}$ & $8^{\text{th}}-10^{\text{th}}$ & $1^{\text{st}}$ & $8^{\text{th}}-10^{\text{th}}$ & $1^{\text{st}}$ & $8^{\text{th}}-10^{\text{th}}$ \\
\midrule
PD-ILC  $[m]$			& $0.213$ & $0.073$ & $0.898$ & $0.120$ & $0.135$ & $0.081$ & $-$ & $-$\\
PID-ILC	$[m]$			& $0.204$ & $0.071$ & $0.745$ & $0.052$ & $0.126$ & $0.049$ & $-$ & $-$\\
$\lone$-ILC $[m]$		& $0.991$ & $0.051$ & $0.044$ & $0.048$ & $0.067$ & $0.050$ & $0.096$ & $0.045$\\
\bottomrule
\end{tabular}
\caption{Average position error of a three-dimensional trajectory flown with the AR.Drone 2.0 for the first and eight to tenth iteration in an ILC experiment for initializing without and with learned experience obtained from (\textit{i}) transferring learning from a dynamically different system, (\textit{ii}) transferring learning from a simulation, and (\textit{iii}) generating the input based on the $\lone$ reference model.}
\label{tbl:comparison}
\end{table}

%% file: Conclusion.tex
In this paper, we show the capabilities of an $\lone$-ILC framework to achieve precise trajectory tracking and to enable transfer learning. The $\lone$ adaptive controller forces the system to remain close to a predefined nominal system behavior, even in the presence of unknown and changing disturbances. This makes it possible for two dynamically different systems to have the same predefined behavior. However, having a repeatable system does not imply achieving zero tracking error. We use ILC to learn from previous iterations and improve the tracking performance over time. We derive performance bounds for the $\lone$-ILC approach analytically. Experimental results on quadrotors show significant performance improvements of the proposed $\lone$-ILC approach compared to a non-adaptive PD-ILC and PID-ILC approach, in terms of disturbance attenuation, transfer learning capability between dynamically different systems, and transfer learning from simulation to the real system. Since the $\lone$ adaptive controller makes the system behave in a predefined way, it also allows us to compute a near optimal input from the $\lone$ reference model, which achieves a small tracking error in the first trial. Overall, the $\lone$-ILC framework promises to make robot learning simpler and more effective as robots can learn from each other and from simulations.

%% file: Appendix.tex
\begin{proof}
\rev{
Substitution of~\eqref{eq:sigmadef} into~\eqref{eq:u1ref}, results in 
\begin{equation}
 u_{1,ref}(s) = \frac{C(s)M(s) r_{1,ref}(s)-C(s)A(s)d_{1,ref}(s)}{C(s) A(s) + (1-C(s))M(s)}\,.
 \label{eq:u1refnosigma}
\end{equation}
From~\eqref{eq:sigmadef} and~\eqref{eq:y1refM}, we obtain,
\begin{equation}
 y_{1,ref}(s) = A(s)(u_{1,ref}(s)+d_{1,ref}(s))\,.
 \label{eq:y1refA}
\end{equation}
Substitution of~\eqref{eq:u1refnosigma} into~\eqref{eq:y1refA} results in 
\[
 y_{1,ref}(s) = A(s)M(s)\frac{C(s)r_{1,ref}(s)+d_{1,ref}(s)(1-C(s))}{C(s)A(s)+(1-C(s))M(s)}\,.
\]
Using~\eqref{eq:defH}, we can rewrite
\begin{equation}
 y_{1,ref}(s) = H(s)(C(s)r_{1,ref}(s)+(1-C(s))d_{1,ref}(s))\,.
 \label{eq:y1refH}
\end{equation}
Substitution of~\eqref{eq:r1ref} into~\eqref{eq:y1refH} and using the definition in~\eqref{eq:y2refs} results in the following expression 
\[
 y_{2,ref}(s) = \frac{1}{s}H(s)(C(s)K(r_{2}(s)-y_{2,ref}(s))+(1-C(s))d_{1,ref}(s))\,,
\]
and hence 
\begin{equation}
 y_{2,ref}(s) = F(s)H(s)(C(s)Kr_2(s)+(1-C(s))d_{1,ref}(s))\,.
 \label{eq:y2refFd}
\end{equation}
In Lemma 4.1.1 in~\cite{Hovakimyan2010}, using the $\lone$-norm condition in~\eqref{eq:l1normcondition}, it is shown that the following upper bound holds uniformly
\[
\lninf{y_{1,ref_\tau}}\leq\frac{\lnone{H(s)C(s)}\lninf{r_{1,ref}}+\lnone{G(s)}L_0}{1-\lnone{G(s)}L}\,,
\]
where $\lninf{y_{1,ref_\tau}}$ is the truncated $\linf$-norm of $y_{1,ref}(t)$ up to $t=\tau$. Hence, $\lninf{y_{1,ref}}$ is bounded. Since $H(s)$, $F(s)$ and $C(s)$ are strictly-proper, stable transfer functions, taking the norm of the reference system~\eqref{eq:y2refFd} and making use of Assumption~\ref{asm:globallip} and Lemma~4.1.1 in~\cite{Hovakimyan2010} yields the following bound: 
\begin{equation}
 \lninf{y_{2,ref_\tau}}\leq\lnone{F(s)H(s)C(s)}K\lninf{r_{2}} + \lnone{F(s)G(s)}(L\lninf{y_{1,ref}}+L_0)\,.
\end{equation}
This results holds uniformly, so $\lninf{y_{2,ref}}$ is bounded. Hence, the closed-loop reference system in (\ref{eq:y1refM}-\ref{eq:r1ref}) is BIBO stable. 
}

\end{proof}

%% file: Appendix2.tex
\begin{proof}
Theorem 4.4.1 in \cite{Hovakimyan2010} proves the bound in \eqref{eq:ytildebound} under the same assumptions made in this paper. It remains to show the bound in \eqref{eq:overallbound}. We use the following definitions: 
\begin{align}
H_0 (s) &\triangleq \dfrac{A(s)}{C(s)A(s) + (1 - C(s)) M(s)} \label{eq:defH0} \ \text{,} \\
H_1 (s) &\triangleq \dfrac{ ( A(s) - M(s) ) C(s) }{C(s)A(s) + (1 - C(s)) M(s)} \ \text{, and}  \label{eq:defH1} \\
\rev{H_2(s)} & \rev{\triangleq \dfrac{C(s)H(s)}{M(s)}\,.}
\end{align}
All $H_0(s)$, $H_1(s)$, and $H_2(s)$ are strictly-proper stable transfer functions, as shown in~\cite{Hovakimyan2010}. The following expressions using~\eqref{eq:defH0} and~\eqref{eq:defH1} can be verified:
\begin{align}
M(s) H_0(s) &= H(s) \ \text{, and} \label{eq:proofexpression1}\\
M(s) \Big( C(s) + &H_1(s) (1 - C(s)) \Big) = H(s) C(s) \,. \label{eq:proofexpression2}
\end{align}
Let $\tilde{\sigma}(t)\triangleq \hat{\sigma}(t)-\sigma(t)$, where $\hat{\sigma}$ is the adaptive estimate and $\sigma$ is defined in \eqref{eq:defsigma}. The control law in \eqref{eq:controllaw} can be expressed as: 
\begin{equation}
u(s) = C(s) r_1(s) - C(s) (\tilde{\sigma}(s) + \sigma(s)) \,. \label{eq:proofcontrol}
\end{equation}
Substitution of~\eqref{eq:proofcontrol} into~\eqref{eq:defsigma} and making use of the definitions in~\eqref{eq:defH0} and~\eqref{eq:defH1} results in the following expression for $\sigma(s)$:
\begin{equation}
\sigma(s) = H_1(s) (r_1(s) - \tilde{\sigma}(s)) + H_0(s) d_\lone(s) \,. \label{eq:proofsigma}
\end{equation}
Substitution of~\eqref{eq:proofcontrol} and~\eqref{eq:proofsigma} into the system~\eqref{eq:newsystem} results in:
\[
y_1(s) = M(s) \Big( C(s) + H_1(s) (1 - C(s)) \Big) \big( r_1(s) - \tilde{\sigma}(s) \big)+ M(s) H_0(s) (1 - C(s)) d_\lone(s) \,. 
\]
Using~\eqref{eq:proofexpression2} and~\eqref{eq:proofexpression1}, this expression simplifies to:
\begin{equation}
y_1(s) = H(s) C(s) \big( r_1(s) - \tilde{\sigma}(s) \big) + H(s) (1 - C(s)) d_\lone(s) \,. \label{eq:finproofsystem}
\end{equation}
We obtain $y_2(s)$ by substituting~\eqref{eq:finproofsystem} and~\eqref{eq:clfeedback} into~\eqref{eq:y2refs} and making use of the definition in~\eqref{eq:defF}:
\begin{equation}
y_2(s) = F(s) H(s) \Big( C(s) K r_2(s) + (1 - C(s)) d_\lone(s) \Big) - F(s) H(s) C(s) \tilde{\sigma}(s)\,. \label{eq:cllonesystem}
\end{equation}
Substituting~\eqref{eq:outputpredictor} and~\eqref{eq:newsystem} into the definition of $\tilde{y}(s)$ in the adaptation law results in:
\begin{equation}
\tilde{y}(s) = M(s) \tilde{\sigma}(s)\,. 
\label{eq:prooftildey}
\end{equation}
Recalling the reference system in~\eqref{eq:y2refFd} and using the expression for $y_2(s)$ in~\eqref{eq:cllonesystem}, the error between reference and actual systems, $y_{2,\text{ref}}(s) - y_2(s)$ is:
\rev{
\[
y_{2,\text{ref}}(s) - y_2(s) = F(s) H(s) \big( 1 - C(s) \big) ( d_{\text{ref}} (s) - d_\lone(s) ) + \dfrac{F(s) H(s) C(s)}{M(s)} M(s) \tilde{\sigma}(s) \,. 
\]
}
Substituting the expression for $\tilde{y}(s)$ in~\eqref{eq:prooftildey} and the definition of G(s) in~\eqref{eq:l1normcondition}, we obtain:
\rev{
\[
y_{2,\text{ref}}(s) - y_2(s) = F(s)G(s) ( d_{\text{ref}} (s) - d_\lone(s) ) 
+\dfrac{F(s) H(s) C(s)}{M(s)} \tilde{y}(s) \,. 
\]
}
\rev{
In Theorem~4.1.1 in~\cite{Hovakimyan2010}, using the same assumptions in this work, the following bound is derived: 
\[
 \lninf{y_{1,ref}-y_1}\leq \frac{\lnone{H_2(s)}}{1-\lnone{G(s)}L}\gamma_0\,.
\]
}
Finally, since the $\lone$-norm of $F(s)G(s)$ exists and $\frac{F(s) H(s) C(s)}{M(s)}$ is strictly-proper and stable, the following bound can be derived by taking the truncated $\linf$-norm and by making use of Assumption~\ref{asm:globallip}: 
\rev{
\[
\lninf{y_{2,{ref}_\tau} - y_{2_\tau}}  \leq  \lnone{F(s)G(s)}L\lninf{ y_{1,ref_\tau} - y_{1_\tau}} + \lnone{\dfrac{F(s) H(s) C(s)}{M(s)} }\lninf{ \tilde{y}_\tau} \,,
\]
which holds uniformly. Making use of the bounds in Theorem~4.1.1 in~\cite{Hovakimyan2010} results in: 
\begin{equation}
 \lninf{y_{2,{ref}} - y_{2}}  \leq \left( \lnone{F(s)G(s)}L\dfrac{\lnone{H_2(s)}}{1-\lnone{G(s)}L} + \lnone{\dfrac{F(s) H(s) C(s)}{M(s)} }\right)\gamma_0 \,,
\end{equation}
proving the bound in~\eqref{eq:overallbound}.
}
\end{proof}

%% file: Appendix3.tex
We begin our discussion with the case where the constraints in \eqref{eq:liftedconst} are inactive. For this case, using \eqref{eq:outputerror} and deriving \eqref{eq:costfunction} with respect to $\bbff{r}{2,j+1}$, we obtain: 
 \[
 \begin{array}{rcl}
 \nabla_{\bar{\bf{r}}_{2,j+1}} & = & (\bff{F}{\text{ILC}}^T\bff{Q}{}\bff{F}{\text{ILC}}+\bff{W}{})\bbff{r}{2,j+1} + \bff{F}{\text{ILC}}^T\bff{Q}{}\hbff{d}{j|j} \,.
 \end{array}
 \]
Equating to 0 and solving for $\bbff{r}{2,j+1}$, we get: 
\begin{equation}
 \bbff{r}{2,j+1} = -(\bff{F}{\text{ILC}}^T \bff{Q}{}\bff{F}{\text{ILC}} + \bff{W}{})^{-1}(\bff{F}{\text{ILC}}^T\bff{Q}{}\hbff{d}{j|j})\,.
 \label{eq:ucinput}
\end{equation}
 By definition $\bff{Q}{}$ is positive definite and $\bff{F}{\text{ILC}}$ is full rank according to Assumption~\ref{ass:Frank}; hence, $\bff{F}{\text{ILC}}^T \bff{Q}{}\bff{F}{\text{ILC}}$ is positive definite. Further, the sum of a positive definite and a positive semidefinite matrix is itself positive definite. Since $\bff{W}{}$ is positive semidefinite by definition $\bff{F}{\text{ILC}}^T \bff{Q}{}\bff{F}{\text{ILC}}+\bff{W}{}$ is positive definite and invertible. The Kalman filter is asymptotically stable; hence, $\hbff{d}{j|j}\rightarrow \bff{d}{\infty}$ as $j\rightarrow\infty$. Therefore, $\bbff{r}{j+1}$ also converges. 
Substituting \eqref{eq:ucinput} into \eqref{eq:lifted}, we  to obtain: 
\begin{equation}
\begin{array}{rcl}
 \bbff{y}{2,j+1} & = & \bff{F}{\text{ILC}} (-(\bff{F}{\text{ILC}}^T \bff{Q}{}\bff{F}{\text{ILC}} + \bff{W}{})^{-1}(\bff{F}{\text{ILC}}^T\bff{Q}{}\hbff{d}{j|j})) + \bff{d}{\infty}\,.
\end{array}
\label{eq:ucoutput}
\end{equation}
Zeroing of the error is possible for the following choice of weighting matrices: $\bff{Q}{}=q\bff{I}{}$ and $\bff{W}{}=0$. Substituting in \eqref{eq:ucoutput}, we obtain:
\[
\begin{array}{rcl}
 \bbff{y}{2,j+1} & = & \bff{F}{\text{ILC}} (-(\bff{F}{\text{ILC}}^T q\bff{I}{}\bff{F}{\text{ILC}} )^{-1}(\bff{F}{\text{ILC}}^T q\bff{I}{}\hbff{d}{j|j})) + \bff{d}{\infty} \\
 & = & -\bff{F}{\text{ILC}} \bff{F}{\text{ILC}}^{-1}\bff{F}{\text{ILC}}^{-T} \bff{F}{\text{ILC}}^T\hbff{d}{j|j} + \bff{d}{\infty}\\
 & = & \bff{d}{\infty} - \hbff{d}{j|j}\,, 
 \end{array}
\]
where $\hbff{d}{j|j}\rightarrow\bff{d}{\infty} $ and $\bbff{y}{2,j+1}\rightarrow 0$ as $j\rightarrow\infty$.

If the inequality constraints are active, we add Lagrangian multipliers to \eqref{eq:costfunction} such that: 
 \[
 \begin{array}{rcl}
  \mathcal{L}(\bar{\bf{r}}_{2,j}, {\bm{\lambda}}_1, {\bm{\lambda}}_2) & = &   \frac{1}{2} \left \{ ({\bf{F}}_{\text{ILC}} \bar{\bf{r}}_{2,j+1} + \widehat{\bf{d}}_{j|j} )^T{\bf{Q}}({\bf{F}}_{\text{ILC}} \bar{\bf{r}}_{2,j+1} + \widehat{\bf{d}}_{j|j} ) + \bar{\bf{r}}_{2,j+1}^T{\bf{W}}\bar{\bf{r}}_{2,j+1}\right\} \\
  & & - \sum_{l\in V_{act}}{\lambda}_{1,l} ( \bff{V}{c,l} \bff{F}{\text{ILC}}\bar{\bf{r}}_{2,j+1} +\bff{V}{c,l} \hbff{d}{j|j} - {\hat{{y}}}_{max} ) \\
  &  & - \sum_{l\in Z_{act}} {\lambda}_{2,l} ( {\bf{Z}}_{c,l} \bar{\bf{r}}_{2,j+1} - \bar{r}_{2,max})  \\
 \end{array}  
 \]
 where $\bf{V}_{c,l}$ is the $l^{th}$ row of $\bff{V}{c}$, $\bf{Z}_{c,l}$ is the $l^{th}$ row of $\bff{Z}{c}$, ${{\lambda}}_{1,l}$ and ${{\lambda}}_{2,l}$ are Lagrange multipliers for the set $V_{act}$ of estimated output $\hbff{y}{j+1|j}$ active constraints and the set $Z_{act}$ of input $\bbff{r}{2,j+1}$ active constraints. The first-order necessary conditions \cite{Wright1999} for $\bbff{r}{2,j+1}$ to be a solution of \eqref{eq:costfunction} subject to \eqref{eq:liftedconst} state that there are vectors $\bm{\lambda}_1^*$ and $\bm{\lambda}_{2}^*$ such that the following system of equations is satisfied: 
 \begin{equation}
 \begin{bmatrix}
 \bff{F}{\text{ILC}}^T \bff{Q}{}\bff{F}{\text{ILC}} + \bff{W}{} & -(\bff{V}{c,act}\bff{F}{\text{ILC}})^T &  -\bff{Z}{c}^T \\
 \bff{V}{c,act}\bff{F}{\text{ILC}} & 0 & 0 \\ 
 \bff{Z}{c,act} & 0 & 0\\
 \end{bmatrix} 
 \begin{bmatrix}
 \bbff{r}{2,j+1} \\
 \bm{\lambda}_1^* \\
 \bm{\lambda}_{2}^*
 \end{bmatrix} = 
 \begin{bmatrix}
 -\bff{F}{\text{ILC}}^T\bff{Q}{}\hbff{d}{j|j} \\
 \hbff{y}{max}-\bff{V}{c,act}\hbff{d}{j|j} \\
 \bff{r}{2,max}
 \end{bmatrix}
 \label{eq:KKT}
 \end{equation}
where $\bff{V}{c,act}$ is the matrix whose rows are $\bff{V}{c,l}\  \forall l \in V_{act}$ and $\bff{Z}{c,act}$ is the matrix whose rows are $\bff{Z}{c,l}\ \forall l \in Z_{act}$. These conditions are a consequence of the first-order optimality conditions described in Theorem 12.2 in \cite{Wright1999}. We denote $L_{V,Z}\leq N$ as the number of elements in $\bff{V}{c,act}\cup \bff{Z}{c,act}$. We use $Z$ to denote the $N\times(N-L_{V,Z})$ matrix whose columns are a basis for the null space of $[\bff{V}{c,act}\bff{F}{\text{ILC}}\ \ \bff{Z}{c,act}]^T$. That is, Z has full rank and satisfies $[\bff{V}{c,act}\bff{F}{\text{ILC}}\ \ \bff{Z}{c,act}]^T Z=0$. 
 
According to Theorem 16.2 in \cite{Wright1999}, if $[\bff{V}{c,act}\bff{F}{\text{ILC}}\ \ \bff{Z}{c,act}]^T$ has full rank and the reduced-Hessian matrix $Z^T(\bff{F}{\text{ILC}}^T \bff{Q}{}\bff{F}{\text{ILC}} + \bff{W}{})Z$ is positive definite, then $\bbff{r}{2,j+1}$ satisfying \eqref{eq:KKT} is the unique global solution of \eqref{eq:costfunction} under \eqref{eq:liftedconst}. We first note that according to Assumption~\ref{ass:inputconstraint}, $[\bff{V}{c,act}\bff{F}{\text{ILC}}\ \ \bff{Z}{c,act}]^T$ has full rank. Since $Z$ has full rank and $\bff{F}{\text{ILC}}^T \bff{Q}{}\bff{F}{\text{ILC}} + \bff{W}{}$ is positive definite as described above, then $Z^T(\bff{F}{\text{ILC}}^T \bff{Q}{}\bff{F}{\text{ILC}} + \bff{W}{})Z$ is positive definite and $\bbff{r}{2,j+1}$ is the unique global solution to the minimization problem. 
 
The unique global solution to the minimization problem with active constraints is:  \begin{equation}
 \bbff{r}{j+1} = (\bff{F}{\text{ILC}}^T \bff{Q}{}\bff{F}{\text{ILC}} + \bff{W}{})^{-1} (-\bff{F}{\text{ILC}}^T\bff{Q}{}\hbff{d}{j|j} + (\bff{V}{c,act}\bff{F}{\text{ILC}})^T\bm{\lambda}_1^* + \bff{Z}{c}^T\bm{\lambda}_2^*)\,.
 \label{eq:cinput}
 \end{equation}
As $j\rightarrow\infty$, $\hbff{d}{j|j}\rightarrow\bff{d}{\infty}$ and $\bbff{r}{j+1}$ converges. Substituting \eqref{eq:cinput} in into \eqref{eq:systemmodel}, we  to obtain: 
\begin{equation}
 \bbff{y}{2,j+1}=\bff{F}{\text{ILC}} (\bff{F}{\text{ILC}}^T \bff{Q}{}\bff{F}{\text{ILC}} + \bff{W}{})^{-1} (-\bff{F}{\text{ILC}}^T\bff{Q}{}\hbff{d}{j|j} + (\bff{V}{c,act}\bff{F}{\text{ILC}})^T\bm{\lambda}_1^* + \bff{Z}{c}^T\bm{\lambda}_2^*) + \bff{d}{\infty}\,.
\label{eq:coutput}
\end{equation}
Zeroing of the error is not possible even with the previous choice of weighting matrices $\bff{Q}{}=q\bff{I}{}$ and $\bff{W}{}=0$:
\[
\begin{array}{rcl}
 \bff{y}{2,j+1} & = & \bff{F}{\text{ILC}} (\bff{F}{\text{ILC}}^T q\bff{I}{}\bff{F}{\text{ILC}} )^{-1} (-\bff{F}{\text{ILC}}^T q\bff{I}{}\hbff{d}{j|j}+ (\bff{V}{c,act}\bff{F}{\text{ILC}})^T\bm{\lambda}_1^* + \bff{Z}{c}^T\bm{\lambda}_2^*) + \bff{d}{\infty} \\
 & = & q^{-1}\bff{F}{\text{ILC}}^{-T}( (\bff{V}{c,act}\bff{F}{\text{ILC}})^T\bm{\lambda}_1^* + \bff{Z}{c}^T\bm{\lambda}_2^*)  + \bff{d}{\infty}-\hbff{d}{j|j}.
 \end{array}
\]

%% file: main.bbl
\begin{thebibliography}{10}
\providecommand{\url}[1]{\texttt{#1}}
\providecommand{\urlprefix}{URL }
\expandafter\ifx\csname urlstyle\endcsname\relax
  \providecommand{\doi}[1]{doi:\discretionary{}{}{}#1}\else
  \providecommand{\doi}{doi:\discretionary{}{}{}\begingroup
  \urlstyle{rm}\Url}\fi

\bibitem{Skelton1989}
Skelton R. Model error concepts in control design. \emph{International Journal
  of Control}  1989; \textbf{49}(5):1725--1753.

\bibitem{Morari1999}
Morari M, Lee JH. Model predictive control: past, present and future.
  \emph{Computers \& Chemical Engineering}  1999; \textbf{23}(4):667--682.

\bibitem{Skogestad2007}
Skogestad S, Postlethwaite I. \emph{Multivariable feedback control: analysis
  and design}, vol.~2. Wiley New York, 2007.

\bibitem{Parks1966}
Parks P. Liapunov redesign of model reference adaptive control systems.
  \emph{IEEE Transactions on Automatic Control}  1966; \textbf{11}(3):362--367.

\bibitem{Hovakimyan2010}
Hovakimyan N, Cao C. \emph{$\lone$ Adaptive Control Theory: Guaranteed
  Robustness with Fast Adaptation}. Society for Industrial and Applied
  Mathematics: Philadelphia, PA, 2010.

\bibitem{Mallikarjunan2012}
Mallikarjunan S, Nesbit B, Kharisov E, Xargay E, Hovakimyan N, Cao C. $\lone$
  adaptive controller for attitude control of multirotors. \emph{Proc. of the
  AIAA Guidance, Navigation and Control Conference}, 2012; 4831.

\bibitem{Michini2009}
Michini B, How JP. $\lone$ adaptive control for indoor autonomous vehicles:
  Design process and flight testing. \emph{Proc. of the AIAA Guidance,
  Navigation and Control Conference}, 2009; 5754.

\bibitem{Gunnarsson2001}
Gunnarsson S, Norrl{\"o}f M. On the design of {ILC} algorithms using
  optimization. \emph{Automatica}  2001; \textbf{37}(12):2011--2016.

\bibitem{Ostafew2013}
Ostafew CJ, Schoellig AP, Barfoot TD. Visual teach and repeat, repeat, repeat:
  Iterative learning control to improve mobile robot path tracking in
  challenging outdoor environments. \emph{Proc. of the IEEE/RSJ International
  Conference on Intelligent Robots and Systems (IROS)}, 2013; 176--181.

\bibitem{Yu2014}
Yu D, Zhu Y, Yang K, Hu C, Li M. A time-varying {Q}-filter design for iterative
  learning control with application to an ultra-precision dual-stage actuated
  wafer stage. \emph{Proc. of the Institution of Mechanical Engineers, Part I:
  Journal of Systems and Control Engineering}  2014; \textbf{228}(9):658--667.

\bibitem{Schoellig2009}
Schoellig AP, D'Andrea R. Optimization-based iterative learning control for
  trajectory tracking. \emph{Proc. of the European Control Conference (ECC)},
  2009; 1505--1510.

\bibitem{Mueller2012}
Mueller FL, Schoellig AP, D'Andrea R. Iterative learning of feed-forward
  corrections for high-performance tracking. \emph{{Proc. of the IEEE/RSJ
  International Conference on Intelligent Robots and Systems (IROS)}}, 2012;
  3276--3281, \doi{10.1109/IROS.2012.6385647}.

\bibitem{Schoellig2012}
Schoellig AP, Mueller FL, D'Andrea R. Optimization-based iterative learning for
  precise quadrocopter trajectory tracking. \emph{Autonomous Robots}  2012;
  \textbf{33}(1-2):103--127.

\bibitem{Bristow2006}
Bristow D, Tharayil M, Alleyne AG. A survey of iterative learning control.
  \emph{IEEE Control Systems}  2006; \textbf{26}(3):96--114.

\bibitem{Barton2011}
Barton K, Mishra S, Xargay E. Robust iterative learning control: $\lone$
  adaptive feedback control in an {ILC} framework. \emph{Proc. of the American
  Control Conference (ACC)}, 2011; 3663--3668.

\bibitem{Altin2013}
Altin B, Barton K. $\lone$ adaptive control in an iterative learning control
  framework: Stability, robustness and design trade-offs. \emph{Proc. of the
  American Control Conference (ACC)}, 2013; 6697--6702.

\bibitem{Altin2014}
Alt{\i}n B, Barton K. Robust iterative learning for high precision motion
  control through $\lone$ adaptive feedback. \emph{Mechatronics}  2014;
  \textbf{24}(6):549--561.

\bibitem{Pereida2017}
Pereida K, Duivenvoorden RR, Schoellig AP. High-precision trajectory tracking
  in changing environments through $\mathcal{L}_1$ adaptive feedback and
  iterative learning. \emph{arXiv preprint arXiv:1705.04763}, 2017.

\bibitem{Tobin2017}
Tobin J, Fong R, Ray A, Schneider J, Zaremba W, Abbeel P. Domain randomization
  for transferring deep neural networks from simulation to the real world.
  \emph{arXiv preprint arXiv:1703.06907}, 2017.

\bibitem{Christiano2016}
Christiano P, Shah Z, Mordatch I, Schneider J, Blackwell T, Tobin J, Abbeel P,
  Zaremba W. Transfer from simulation to real world through learning deep
  inverse dynamics model. \emph{arXiv preprint arXiv:1610.03518}, 2016.

\bibitem{Devin2016}
Devin C, Gupta A, Darrell T, Abbeel P, Levine S. Learning modular neural
  network policies for multi-task and multi-robot transfer. \emph{arXiv
  preprint arXiv:1609.07088}, 2016.

\bibitem{Hamer2013}
Hamer M, Waibel M, D'Andrea R. Knowledge transfer for high-performance
  quadrocopter maneuvers. \emph{Proc. of the IEEE/RSJ International Conference
  on Intelligent Robots and Systems (IROS)}, 2013; 1714--1719.

\bibitem{Morari1985}
Morari M. Robust stability of systems with integral control. \emph{IEEE
  Transactions on Automatic Control}  1985; \textbf{30}(6):574--577.

\bibitem{Konstantopoulos2016}
Konstantopoulos GC, Zhong QC, Ren B, Krstic M. Bounded integral control of
  input-to-state practically stable nonlinear systems to guarantee closed-loop
  stability. \emph{IEEE Transactions on Automatic Control}  2016;
  \textbf{61}(12):4196--4202.

\bibitem{Lee2000}
Lee JH, Lee KS, Kim WC. Model-based iterative learning control with a quadratic
  criterion for time-varying linear systems. \emph{Automatica}  2000;
  \textbf{36}(5):641--657.

\bibitem{Powers2015}
Powers C, Mellinger D, Kumar V. Quadrotor kinematics and dynamics.
  \emph{Handbook of Unmanned Aerial Vehicles}. Springer, 2015; 307--328.

\bibitem{schoellig2010}
Schoellig AP, Augugliaro F, D'Andrea R. Synchronizing the motion of a
  quadrocopter to music. \emph{{Proc. of the IEEE International Conference on
  Robotics and Automation (ICRA)}}, 2010; 3355--3360.

\bibitem{Wright1999}
Wright S, Nocedal J. Numerical optimization. \emph{Springer Science}  1999;
  \textbf{35}:67--68.

\end{thebibliography}
